%% file: main.tex
\documentclass[10pt,twocolumn,letterpaper]{article}
\usepackage[pagenumbers]{cvpr} 
\usepackage[accsupp]{axessibility}

\usepackage{marvosym}
\usepackage{xcolor}
\usepackage{colortbl}
\usepackage{multirow}
\usepackage{array}
\usepackage{siunitx}
\usepackage{lipsum}
\usepackage{bm}
\usepackage{amsmath}
\usepackage{amsfonts}
\RequirePackage{makecell}

\newcommand{\psiv}{\boldsymbol{\psi}}
\newcommand{\thetav}{\boldsymbol{\theta}}
\newcommand{\phiv}{\boldsymbol{\phi}}

\definecolor{cvprblue}{rgb}{0.21,0.49,0.74}
\usepackage[pagebackref,breaklinks,colorlinks,allcolors=cvprblue]{hyperref}

\def\paperID{13171} 
\def\confName{CVPR}
\def\confYear{2025}

\title{RoboBrain: A Unified Brain Model for Robotic Manipulation \\from Abstract to Concrete}

\author{\small Yuheng Ji$^{2,3,6,*}$, Huajie Tan$^{1,2,*}$,  Jiayu Shi$^{1,2,*}$, Xiaoshuai Hao$^{2,*,\dagger}$, Yuan Zhang$^{1,2}$, Hengyuan Zhang$^{1,2}$\\
\small Pengwei Wang$^{2,\dagger}$, Mengdi Zhao$^2$, Yao Mu$^5$, Pengju An$^{1,2}$, Xinda Xue$^{1,2}$, Qinghang Su$^{2,4}$, Huaihai Lyu$^{2,3,6}$ \\
\small Xiaolong Zheng$^{3,6}$, Jiaming Liu$^{1,2}$, Zhongyuan Wang$^2$, Shanghang Zhang$^{1,2,\text{\Letter}}$ \\
$^1$ \small State Key Laboratory of Multimedia Information Processing, School of Computer Science, Peking University \\ 
$^2$ \small Beijing Academy of Artificial Intelligence
$^3$ \small Institute of Automation, Chinese Academy of Sciences \\
$^4$ \small Institute of Information Engineering, Chinese Academy of Sciences 
$^5$ \small The University of Hong Kong\\
$^6$ \small School of Artificial Intelligence, University of Chinese Academy of Sciences
}

\begin{document}
\twocolumn[{%
\maketitle
\vspace{-0.9cm}
\begin{center}
    \centering
    \captionsetup{type=figure}
    \includegraphics[width=0.88\linewidth]{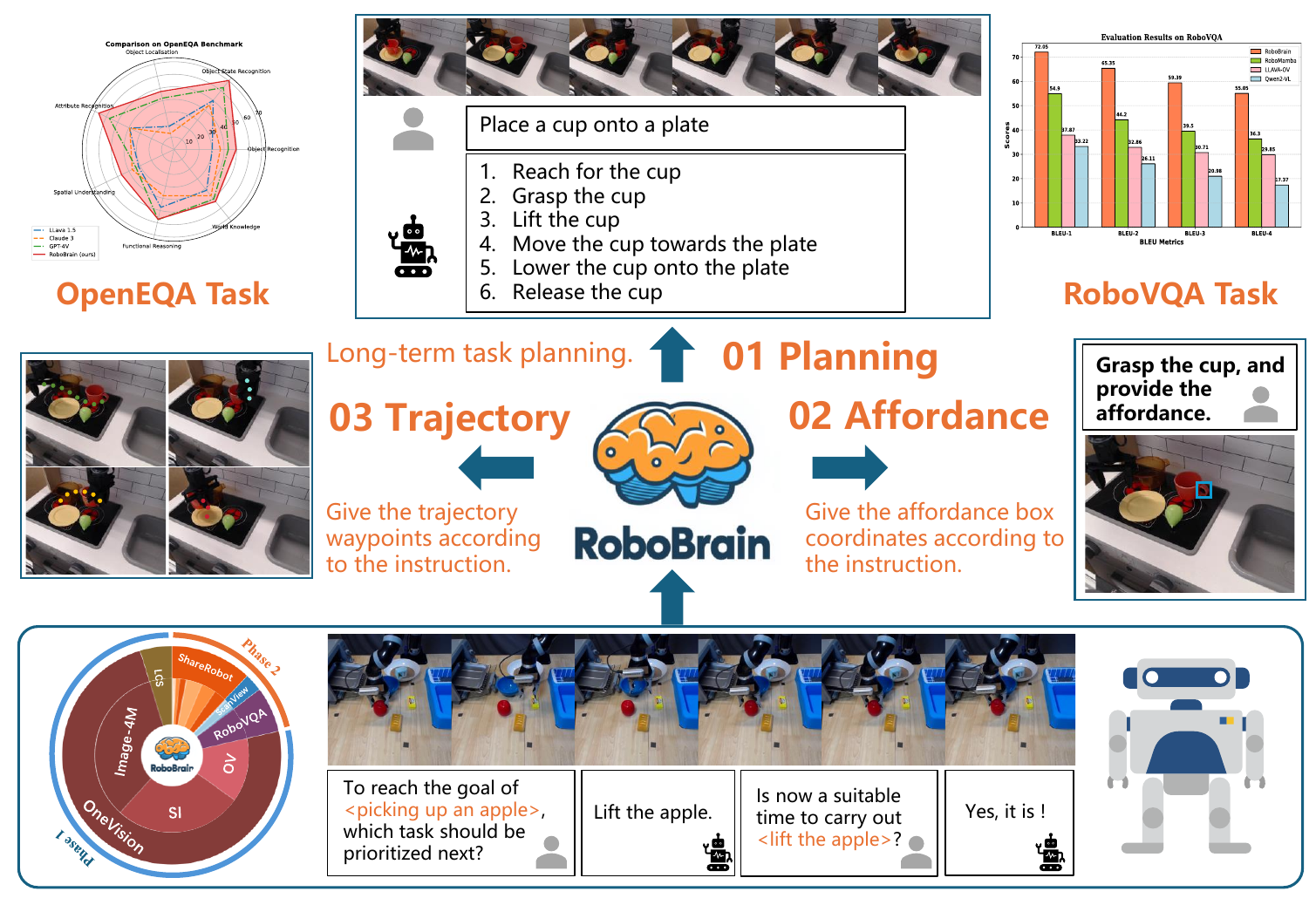}
    \captionof{figure}{
    \textbf{Overview of RoboBrain.} RoboBrain consists of three key robotic capabilities: planning capability, affordance perception, and trajectory prediction.
    RoboBrain outperforms previous MLLMs in robotics tasks.
    The bottom part shows the composition of RoboBrain's training data and provides a specific example of visual question answering from our proposed ShareRobot.
    Best viewed on screen.
    } 
    \label{fig:teaser}
\end{center}
}]

\let\thefootnote\relax\footnotetext{$^{*}$ Equal contribution.}
\let\thefootnote\relax\footnotetext{$^{\dagger}$ Project leaders.}
\let\thefootnote\relax\footnotetext{$^{\text{\Letter}}$ Corresponding author.}

\input{sec/0_abstract}

\input{sec/1_introduction}
\input{sec/2_related_work}
\input{sec/3_1_data}
\input{sec/3_2_model}
\input{sec/4_experiment}
\input{sec/6_conclusion}
{
    \small
    \bibliographystyle{ieeenat_fullname}
    \bibliography{main}
}

\clearpage
\input{sec/X_suppl}

\end{document}

%% file: sec/0_abstract.tex
\begin{abstract}

Recent advancements in Multimodal Large Language Models (MLLMs) have shown remarkable capabilities across various multimodal contexts. 
However, their application in robotic scenarios, particularly for long-horizon manipulation tasks, reveals significant limitations. 
These limitations arise from the current MLLMs lacking three essential robotic brain capabilities:
\textbf{Planning Capability}, which involves decomposing complex manipulation instructions into manageable sub-tasks; \textbf{Affordance Perception}, the ability to recognize and interpret the affordances of interactive objects; and \textbf{Trajectory Prediction}, the foresight to anticipate the complete manipulation trajectory necessary for successful execution.
To enhance the robotic brain's core capabilities from abstract to concrete, we introduce \textbf{ShareRobot}, a high-quality heterogeneous dataset that labels multi-dimensional information such as task planning, object affordance, and end-effector trajectory.
ShareRobot's diversity and accuracy have been meticulously refined by three human annotators.
Building on this dataset, we developed \textbf{RoboBrain}, an MLLM-based model that combines robotic and general multi-modal data, utilizes a multi-stage training strategy, and incorporates long videos and high-resolution images to improve its robotic manipulation capabilities.
Extensive experiments demonstrate that RoboBrain achieves state-of-the-art performance across various robotic tasks, highlighting its potential to advance robotic brain capabilities. Project website: \href{https://superrobobrain.github.io/}{RoboBrain}.

\end{abstract}

%% file: sec/1_introduction.tex
\vspace{-1em}
\section{Introduction} 

Recent advancements in Multimodal Large Language Models (MLLMs) have significantly advanced the pursuit of Artificial General Intelligence (AGI). 
By leveraging extensive multimodal datasets sourced from the internet and employing self-supervised learning techniques, MLLMs demonstrate exceptional capabilities in visual perception and understanding human language instructions, excelling in tasks such as visual question answering~\cite{vqa,pali,internvl}, image captioning~\cite{mplug,hao2023mixgen,blip2}, and sentiment analysis~\cite{sentiment,du2024financial}.
Despite significant progress in MLLMs, the exploration of their application in robotics remains in its early stages, highlighting a crucial area for further research and innovation.

Recent studies have examined the application of MLLMs in robotics, focusing on planning and subgoal decomposition~\cite{Rt-h,inner}, action sequencing~\cite{Saycan,Rt-2}, and replanning and feedback~\cite{reflect,selfcorrect,code_as_monitor}. However, their effectiveness in robotic scenarios—particularly for long-horizon manipulation tasks—reveals significant limitations. These limitations stem from the current MLLMs' lack of three critical robotic capabilities: planning, affordance perception, and trajectory prediction, as illustrated in Fig.~\ref{fig:teaser}.
For instance, consider a robotic arm tasked with lifting a teapot and pouring water into a cup. The MLLM should be capable of decomposing this task into sub-tasks, such as ``approach the teapot and lift it'', ``move the teapot until the spout is positioned over the cup'', and ``tilt the teapot to pour''. For each sub-task, such as ``approach and grasp the teapot'', the MLLM must utilize affordance perception to accurately identify the graspable regions of the teapot. Additionally, trajectory prediction is essential for determining the complete path from the starting point to the graspable part of the teapot.
This challenge for existing MLLMs primarily arises from the scarcity of large-scale, fine-grained datasets specifically designed for robotic operation tasks.

To empower the RoboBrain’s core capabilities that transition from abstract instruction comprehension to concrete action expression.
we first introduce \textbf{\textit{ShareRobot}}, a large-scale, fine-grained dataset specifically designed for robotic operation tasks.
Specifically, we label multi-dimensional information such as task planning, object affordance, and end-effector trajectory.
Building upon ShareRobot, we developed \textbf{\textit{RoboBrain}}, an MLLM model based on the LLaVA~\cite{LLaVa} architecture, aimed at enhancing the perception and planning capabilities of robots in complex tasks. 
In the process of training RoboBrain, we meticulously designed the ratio of robotic data to general multi-modal data, implemented a multi-stage training strategy, and incorporated long videos and high-resolution images. 
This approach endowed RoboBrain with powerful visual information perception capabilities in robotic scenarios, supporting historical frame memory and high-definition image input, thereby further enhancing the ability in robotic manipulation planning. 
Extensive experimental results demonstrate that RoboBrain outperforms existing models across multiple robotic benchmarks, including RoboVQA~\cite{RoboVQA} and OpenEQA~\cite{OpenEQA}, achieving state-of-the-art performance. 
Additionally, it shows competitive results in trajectory and affordance prediction accuracy. 
These findings validate the effectiveness of the proposed dataset and framework in enhancing robotic brain capabilities.
In summary, the main contributions of this paper are as follows:
\begin{itemize}
    \item
    We propose \textbf{\textit{RoboBrain}}, a unified multimodal large language model designed for robotic manipulation, which facilitates more efficient task execution by transforming abstract instruction into concrete actions.
    \item We meticulously designed the ratio of robotic data to general multi-modal data, implemented a multi-stage training strategy, and incorporated long videos and high-resolution images. This approach provided \textbf{\textit{RoboBrain}} with historical frame memory and high-resolution image input, thereby further enhancing its capabilities in robotic manipulation planning.
    \item 
    We introduce \textbf{\textit{ShareRobot}}, a high-quality heterogeneous dataset that labels multi-dimensional information, including task planning, object affordance, and end-effector trajectory, effectively enhancing various robotic capabilities.
    \item 
    Comprehensive experimental results demonstrate that \textbf{\textit{RoboBrain}} achieves state-of-the-art performance across various robotic benchmarks, highlighting its potential for real-world applications in robotics.
\end{itemize}

%% file: sec/2_related_work.tex
\begin{figure*}[!h]
    \centering
    \includegraphics[width=0.95\linewidth]{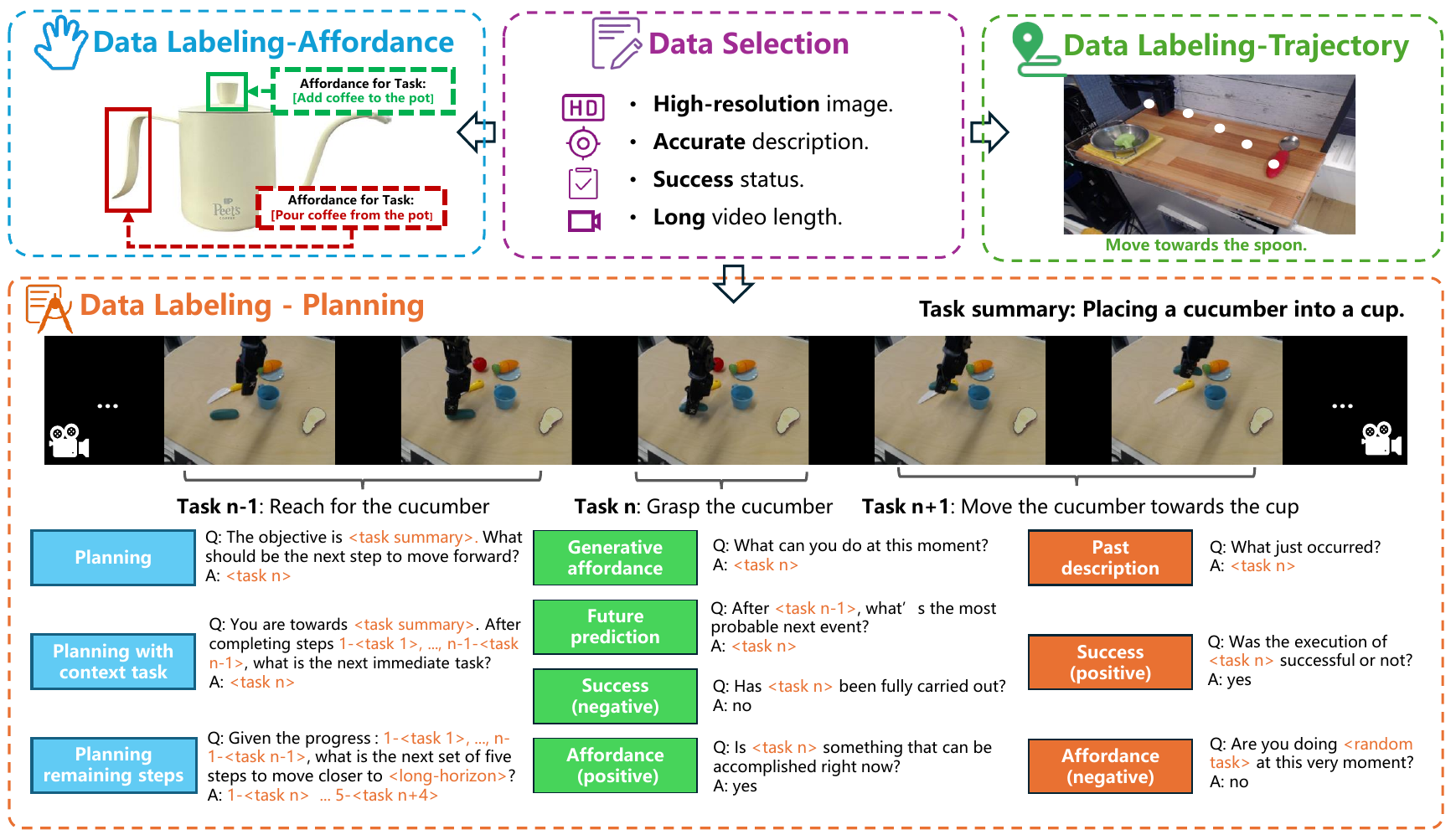}
    \caption{\textbf{The generation procession of our ShareRobot dataset.} Our dataset labels multi-dimensional information, including task planning, object affordance, and end-effector trajectories. The task planning is first annotated by atomic tasks and then augmented by constructing question-answer pairs. The affordance and trajectory are labeled on the images according to the specific instructions.}
    \label{fig:pipeline}
\end{figure*}

\section{Related Work}
\textbf{MLLM for Robotic Manipulation Planning}
Existing studies mostly utilize MLLMs primarily focus on understanding natural language and visual observation tasks~\cite{Rt-1,Rt-2,Rt-h,OpenVLA,li2024foundation,zhang2025mapnav}, with fewer addressing the decomposition of high-level task instructions into actionable steps.
PaLM-E~\cite{Palm-e} generates multimodal inputs by mapping real-world observations into the language embedding space. RT-H~\cite{Rt-h} and RoboMamba~\cite{Robomamba} generate reasoning results along with robot actions obtained from an additional policy head. However, while these models generate planning texts and actions, they still lack adequate mechanisms for executing complex atomic tasks, highlighting the need for enhanced affordance perception and trajectory prediction.

\textbf{Datasets for Manipulation Planning}
Early datasets for Manipulation~\cite{DexYCB,HOnnotate,H2O,TACO,GRAB} mainly comprise annotated images and videos that highlight fundamental hand-object interactions, including grasping and pushing.
Recent advancements~\cite{RoboNet, RoboVQA,tang2025affordgrasp,hao2025tla} in robotic manipulation emphasize multi-modal and cross-embodiment datasets for enhanced generalization. 
Datasets such as RH20T~\cite{RH20T}, BridgeDataV2~\cite{BridgeDatav2}, and DROID~\cite{Droid} enhance scene diversity, broadening the range of manipulation scenarios. 
Notably, RT-X~\cite{RT-X} compiles data from 60 datasets across 22 embodiments into the Open X-Embodiment (OXE) repository.
In this work, we extract high-quality data from OXE, decompose high-level descriptions into low-level planning instructions, and adapt these into a question-answer format to enhance model training.

%% file: sec/3_1_data.tex
\begin{figure*}[!h]
    \centering
    \includegraphics[width=1.0\linewidth]{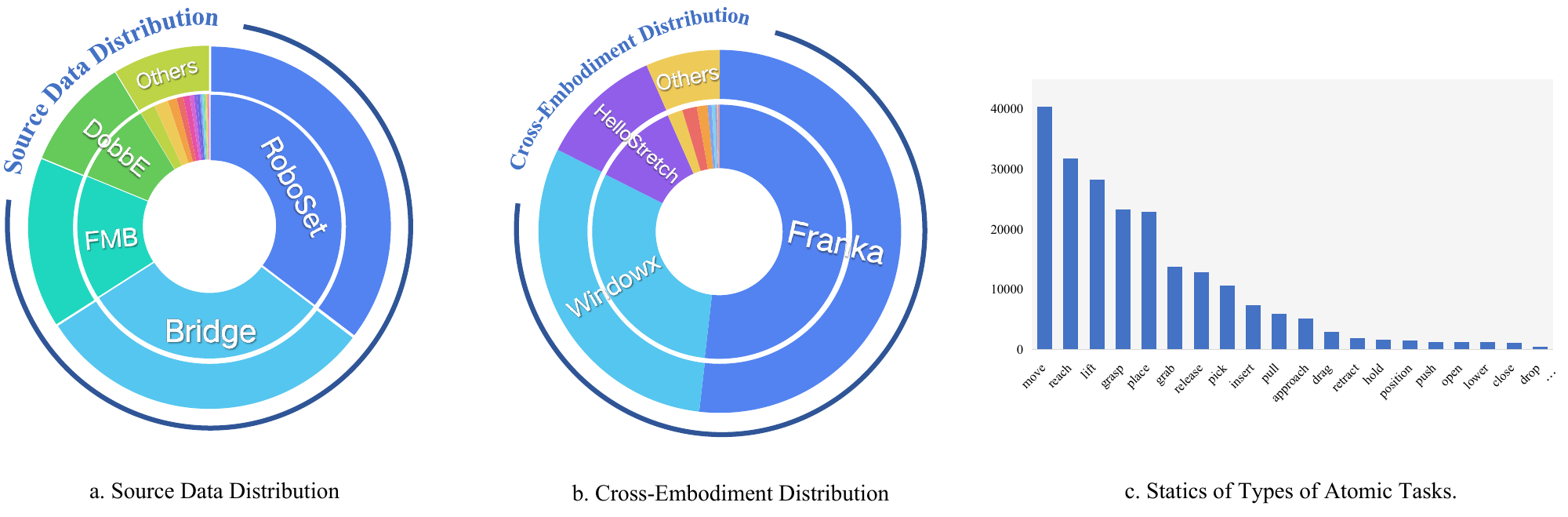}
    \setlength{\abovecaptionskip}{-1em}
    \caption{\textbf{The diversity of our ShareRobot dataset.} Our dataset involves (a) 23 original datasets, (b) 12 embodiments and (c) 107 types of atomic tasks. The distribution of the top 20 most frequent atomic actions within our ShareRobot dataset is presented in (c).}
    \label{fig:mul-dis}
\end{figure*}

\section{ShareRobot Dataset}\label{sec:shareRobot}
To enhance the RoboBrain's capability of planning, affordance perception, and trajectory prediction, we develop a dataset called ShareRobot--a large-scale, fine-grained dataset specifically designed for robotic manipulation tasks. The generation procession of our dataset is shown as Fig.~\ref{fig:pipeline}. The details are described in the following sections.

\subsection{Overview}
ShareRobot is a comprehensive dataset, facilitates more efficient task execution by transforming abstract concepts into concrete actions. The main features of the \textbf{\textit{ShareRobot}} dataset include:

\begin{itemize}
\item \textbf{Fine-grained} Unlike the Open X-Embodiment dataset~\cite{o2023open}, which provides generalized high-level task descriptions, each data point in ShareRobot includes detailed low-level planning instructions linked to individual frames. This specificity enhances the model's accuracy in executing tasks at the right moment.
\item \textbf{Multi-dimensional} To enhance RoboBrain's capabilities from abstract to concrete, we label task planning, object affordances, and end-effector trajectories, allowing for greater flexibility and precision in task processing.
\item \textbf{High quality} We establish rigorous criteria for selecting data from the Open-X-Embodiment dataset~\cite{o2023open}, focusing on high resolution, accurate descriptions, successful task execution, visible affordance, and clear motion trajectories. Based on these criteria, we validate 51,403 instances to ensure high quality, forming the foundation for RoboBrain's core capabilities.
\item \textbf{Large scale} With 1,027,990 question-answer pairs, ShareRobot is the largest open-source dataset for task planning, affordance perception, and trajectory prediction, enabling deeper understanding of complex relationships from abstract to concrete.
\item \textbf{Rich diversity} In contrast to the RoboVQA~\cite{RoboVQA} dataset's limited scenes, ShareRobot features 102 scenes across 12 embodiments and 107 types of atomic tasks, as shown in Fig.~\ref{fig:mul-dis}. This diversity allows MLLMs to learn from varied real-world contexts, enhancing robustness in complex, multi-step planning.
\item \textbf{Easy scalability} Our data generation pipeline is designed for high scalability, facilitating expansion as new robotic embodiments, task types, and environments develop. This adaptability ensures the ShareRobot dataset can support increasingly complex manipulation tasks.
\end{itemize}

\begin{figure*}[!h]
    \centering
    \includegraphics[width=1\linewidth]{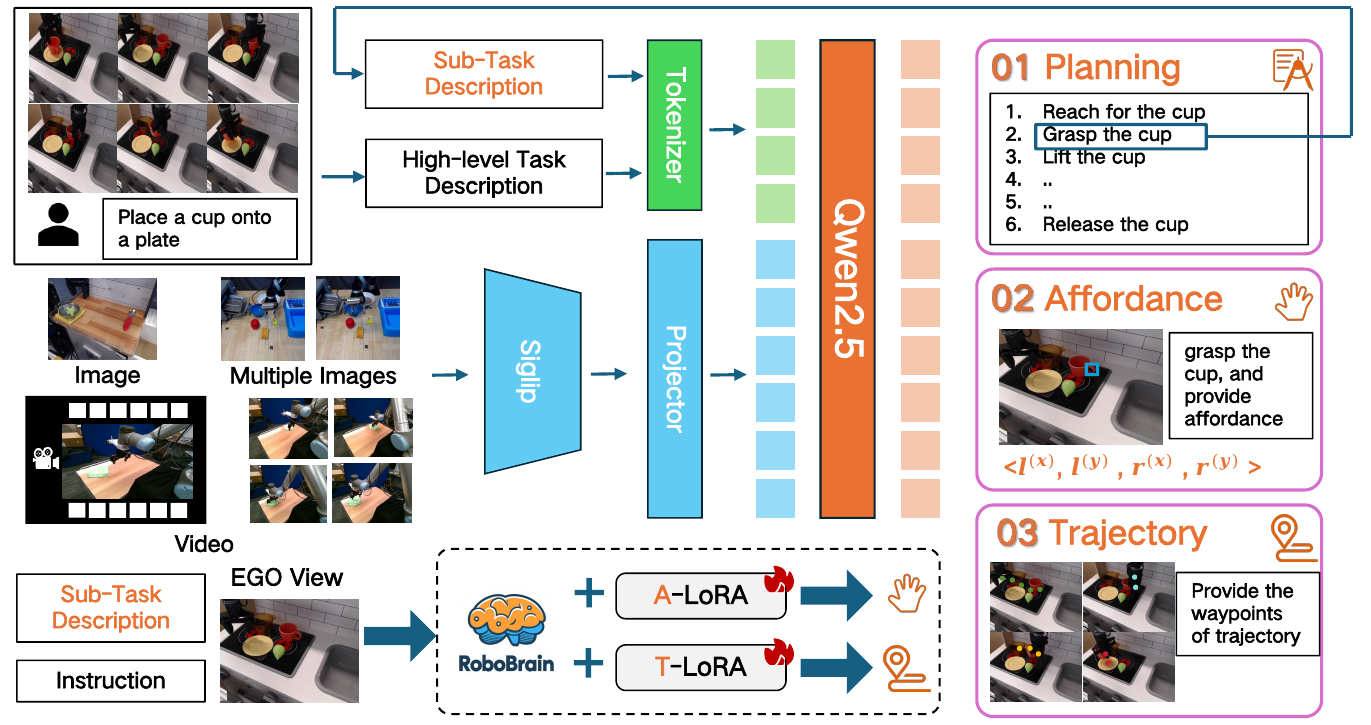}
    \caption{\textbf{The pipeline of our RoboBrain.} 
    The images, multiple images, and videos are sent into our model to pre-train a foundation robotic brain. 
    Besides, we fine-tune the RoboBrain via A-LoRA and T-LoRA to develop affordance and trajectory skills. 
    In practical applications, the model first generates detailed plans, and then splits it into sub-task descriptions to execute specific robotic tasks. }
    \label{fig:model}
\vspace{-0.5em}
\end{figure*}

\subsection{Data Selection}

Based on the Open X-embodiment dataset~\cite{o2023open}, we carefully selected 51,403 instances, mainly focusing on image quality, description accuracy and success status. Our data collection process adheres to the following principles:

\begin{itemize}
    \item \textbf{High-resolution image} We eliminate videos lacking images or those with low resolution. Any video with a resolution below 128 pixels is removed.
    
    \item \textbf{Accurate description} Videos without descriptions or with vague descriptions are filtered out to avoid affecting the planning capability of the model.

    \item \textbf{Success status} We discard videos of failed tasks, as unsuccessful demonstrations hinder the model's learning.
    
    \item \textbf{Long video length} Videos with fewer than 30 frames are excluded, as they contain limited atomic tasks.  

    \item \textbf{Object not covered} We remove any videos where the target object or end-effector is covered by other objects, as our model has to accurately identify the positions of end-effectors and the object's affordance.
   
    \item \textbf{Clear Trajectories} We exclude the demonstrations with unclear or incomplete trajectories, as trajectory prediction is one of our RoboBrain's capabilities.
\end{itemize}

\subsection{Data Labeling}\label{sec:data-labeling}
\textbf{Planning Labeling}
We extract 30 frames from each robotic operation demonstration and use these frames along with their high-level descriptions to decompose them into low-level planning instructions using Gemini~\cite{geminiteam2024geminifamilyhighlycapable}. Three annotators then review and refine these instructions to ensure the precision of labeling. Subsequently, we design 5 different templates for each of the 10 question types in RoboVQA~\cite{RoboVQA}. In the process of data generation, we randomly select 2 templates of each question type to generate question-answer pairs for every instance. This process transforms 51,403 instances into 1,027,990 question-answer pairs, with annotators monitoring data generation to maintain the dataset's integrity.

\textbf{Affordance Labeling}
We filter 6,522 images and annotate each with affordance areas as \{${l^{(x)}, l^{(y)},  r^{(x)}, r^{(y)}}$\} according to its high-level description, where \{${l^{(x)}, l^{(y)}}$\} are the top left coordinates and $\{r^{(x)}, r^{(y)}\}$ are the bottom right corner coordinates. Subsequently, we conduct a rigorous manual review and refinement of each instruction to ensure its precise alignment with the associated affordance areas.

\textbf{Trajectory Labeling}
We filter 6,870 images and annotate each with the gripper's trajectory using at least three $\{x,y\}$ coordinates according to its low-level instruction. Subsequently, we conduct a rigorous manual review and refinement of each instruction to ensure its precise alignment with the associated trajectory.

\subsection{Data Statistics}

We select 23 original datasets from the Open X-embodiment dataset~\cite{o2023open}. The distribution of the source data is shown in the Fig.~\ref{fig:mul-dis}. The data involves 102 various scenes (e.g. bedroom, laboratory, kitchen, office), and covers 12 different robot bodies. According to statistics, there are 132 types of atomic actions in this dataset, tasks with higher word frequency are shown in Fig.~\ref{fig:mul-dis} (c). The 5 most frequent atomic tasks are ``pick'', ``move'', ``reach'', ``lift'', and ``place'', which are frequent task types in real robotic operation scenarios. This suggests that the distribution of our dataset is reasonable. Finally, we get 1,027,990 question-answer (QA) pairs for planning. For the planning QA pairs dataset, we split 1 million QA pairs as the training set and 2,050 QA pairs as the test set. For the affordance dataset, we split 6,000 images as the training set and 522 images as the test set. For the trajectory dataset, we split 6000 images for training and 870 images for testing.

%% file: sec/3_2_model.tex

\begin{table*}[t]
    \centering
    
    \includegraphics[width=0.9\textwidth]{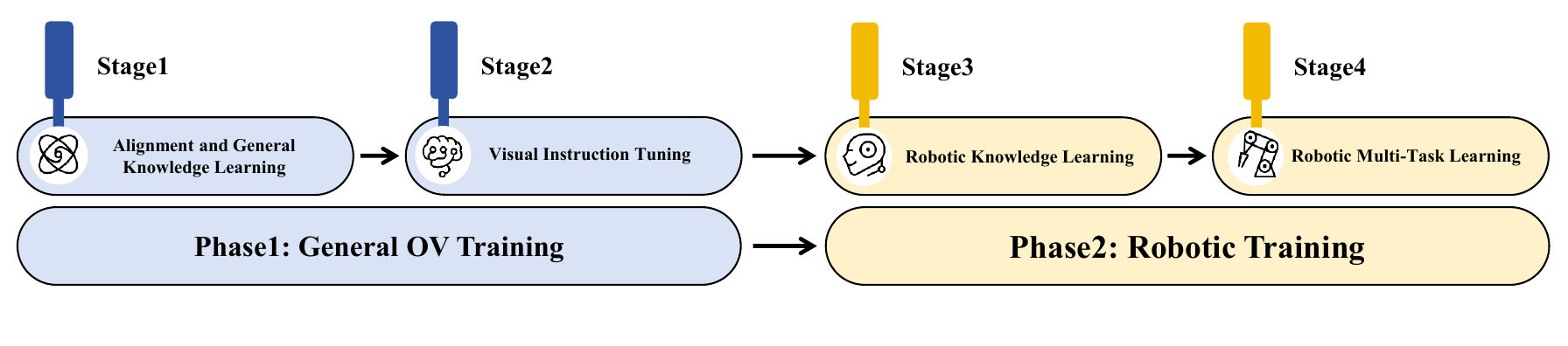}
    \hfill
    \setlength{\tabcolsep}{12pt}
    \renewcommand{\arraystretch}{1.2}
    \resizebox{\textwidth}{!}{%
    \begin{tabular}{@{}ll|c|c|c|c|c|c|c@{}}
        \toprule
        & & \textbf{Stage-1} & \textbf{Stage-1.5} & \multicolumn{2}{c|}{\textbf{Stage-2}} & \textbf{Stage-3} & \multicolumn{2}{c}{\textbf{Stage-4}} \\ \cmidrule(l){5-6} \cmidrule(l){8-9}
        & & & & \textbf{Single-Image} & \textbf{OneVision} &  & \textbf{A-LoRA} & \textbf{T-LoRA} \\
        \midrule 
        \multirow{2}{*}{\rotatebox[origin=c]{90}{\small \textit{Vision}}}
        & \textbf{Resolution}  & 384 & Max 384\footnotesize{$\times$\{2$\times$2\}} & Max 384\footnotesize{$\times$\{6$\times$6\}} & Max 384\footnotesize{$\times$\{6$\times$6\}} & Max 384\footnotesize{$\times$\{6$\times$6\}} & Max 384\footnotesize{$\times$\{6$\times$6\}} & Max 384\footnotesize{$\times$\{6$\times$6\}} \\
        & \#Tokens & 729 & Max 729\footnotesize{$\times$5} & Max 729\footnotesize{$\times$37} & Max 729{\footnotesize{$\times$37}} & Max 729{\footnotesize{$\times$37}} & Max 729{\footnotesize{$\times$37}} & Max 729{\footnotesize{$\times$37}} \\
        \midrule 
        \multirow{2}{*}{\rotatebox[origin=c]{90}{\small \textit{Data}}}
        & \textbf{Dataset} & LCS & Image & Image & Image \& Video & Robotic Data & Afford. Data & Traj. Data \\
        & \#Samples & 558K & 4M & 3.2M & 1.6M & 3M & 10K & 400K \\
        \midrule
        \multirow{2}{*}{\rotatebox[origin=c]{90}{\small \textit{Model}}}
        & \textbf{Trainable} & Projector & Full Model & Full Model & Full Model & Full Model & A-LoRA & T-LoRA \\
        & \#Tunable Parameters & 17.0M & 8.0B & 8.0B & 8.0B & 8.0B & 28.0M & 28.0M \\
        \midrule 
        \multirow{4}{*}{\rotatebox[origin=c]{90}{\small \textit{Training}}}
        & \textbf{Batch Size} & 8 & 2 & 1 & 1 & 1 & 4 & 4 \\
        & \textbf{LR: $\psiv_{\text{ViT}}$} & - & 2 $\times 10^{-6}$ & 2 $\times 10^{-6}$ & 2 $\times 10^{-6}$ & 2 $\times 10^{-6}$ & 2 $\times 10^{-6}$ & 2 $\times 10^{-6}$ \\    
        & \textbf{LR: $\{\thetav_{\text{Proj.}}, \phiv_{\text{LLM}}, \phiv_{\text{LoRA}}\}$} & 1$\times 10^{-3}$ & 1 $\times 10^{-5}$ & 1 $\times 10^{-5}$ & 1 $\times 10^{-5}$ & 1 $\times 10^{-5}$ & 1 $\times 10^{-5}$ & 1 $\times 10^{-5}$ \\
        & \textbf{Epoch} & 1 & 1 & 1 & 1 & 1 & 1 & 1 \\
        \bottomrule
    \end{tabular}
    }
    \vspace{1mm}
    \caption{Detailed configuration for each training stage of the RoboBrain.}
    \vspace{-0.5em}
    \label{tab:training_strategy}
\end{table*}
\section{RoboBrain Model}
In this section, we provide an overview of \textbf{\textit{RoboBrain}}. Our goal is to enable the Multi-modal Large Language Model (MLLM) to understand abstract instructions and explicitly output object affordance regions and potential operational trajectories, facilitating a transition from abstract to concrete.
We employ a multi-stage training strategy: Phase 1 focuses on general OneVision (OV) training to develop a foundational MLLM with strong understanding and instruction-following abilities. 
Phase 2, the robotic training phase, aims to empower the core capabilities of RoboBrain from abstract to concrete.

\subsection{Model Architecture}
RoboBrain consists of three modules: the foundational model for planning, the A-LoRA model for affordance perception, and the T-LoRA model for trajectory prediction. In practical applications, the model first generates detailed plans, and then splits it into sub-task descriptions to execute affordance perception and trajectory prediction. The pipeline of our RoboBrain is shown to Fig.~\ref{fig:model}.

\textbf{Foundational Model for Planning}
We utilize LLaVA as the foundational model for RoboBrain, which consists of three main modules: the Vision Encoder (ViT) $g(\cdot)$, the Projectior $h(\cdot)$, and the Large Language Model (LLM) $f(\cdot)$. Specifically, we employ SigLIP~\cite{siglip}, a 2-layer MLP~\cite{mlp}, and Qwen2.5-7B-Instruct~\cite{qwen2_5}. Given an image or video $X_v$ as visual input, ViT encodes it into visual features $Z_v = g(X_v)$, which are then mapped to the semantic space of the LLM through Projector, resulting in a sequence of visual tokens $H_v = h(Z_v)$. Finally, the LLM generates a textual response in an autoregressive manner based on the human language instruction $X_t$ and $H_v$.

\textbf{A-LoRA Module for Affordance Perception}
The term affordance in our work refers to the area where the human hand makes contact with objects. During interactions, humans instinctively engage with various objects within specific regions. We utilize \textit{bounding boxes} to represent affordances. Formally, consider an image $I$ consisting of multiple objects with their affordances: $O_i = \{A_{i}^{0}, A_{i}^{1}, ..., A_{i}^{N}\}$, where the $i$th object owns $N$ affordances. The format of affordance is defined as $\{l^{(x)}, l^{(y)}, r^{(x)}, r^{(y)}\}$, and $\{l^{(x)}, l^{(y)}\}$ represents the top left corner coordinates of affordance, while $\{r^{(x)}, r^{(y)}\}$ is the bottom right corner coordinates.

\textbf{T-LoRA Module for Trajectory Prediction}
The term trajectory in our work refers to the concept of \textit{2D visual traces}, as presented in \cite{gu2023rt}. We define trajectory waypoints as a series of 2D coordinates representing the movement of the end-effector or hand throughout the process. Formally, at time step $t$, the trajectory waypoints can be represented as $P_{t:N} = \{(x_i, y_i) \mid i = t, t + 1, \dots, N\}$, where $(x_i, y_i)$ denotes the $i$-th coordinate in the visual trace, and $N$ represents the total number of time steps in the episode.

\subsection{Training}
\textbf{Phase 1: General OV Training}
In Phase 1, we drew on the state-of-the-art training data and strategies from LLaVA-OneVision~\cite{li2024llava} to construct a foundational model with general multi-modal understanding and visual instruction following capabilities. This lays the groundwork for enhancing the model's robotic manipulation planning abilities in Phase 2. Detailed information is provided in Tab.~\ref{tab:training_strategy}.

In Stage 1, we utilize the image-text data from the LCS-558K dataset~\cite{cc3m, laion} to train Projector, facilitating the alignment of visual features \( Z_v \) with the LLM semantic features \( H_v \). In Stage 1.5, we train the entire model using 4M high-quality image-text data to enhance the model's multi-modal general knowledge understanding capabilities. In Stage 2, we further train the entire model with 3.2M single-image data and 1.6M image and video data from LLaVA-OneVision-Data~\cite{li2024llava}, aiming to enhance the instruction-following abilities of RoboBrain and improve understanding of high-resolution image and video.

\begin{figure*}[!h]
    \raggedright
    \includegraphics[width=0.98\linewidth]{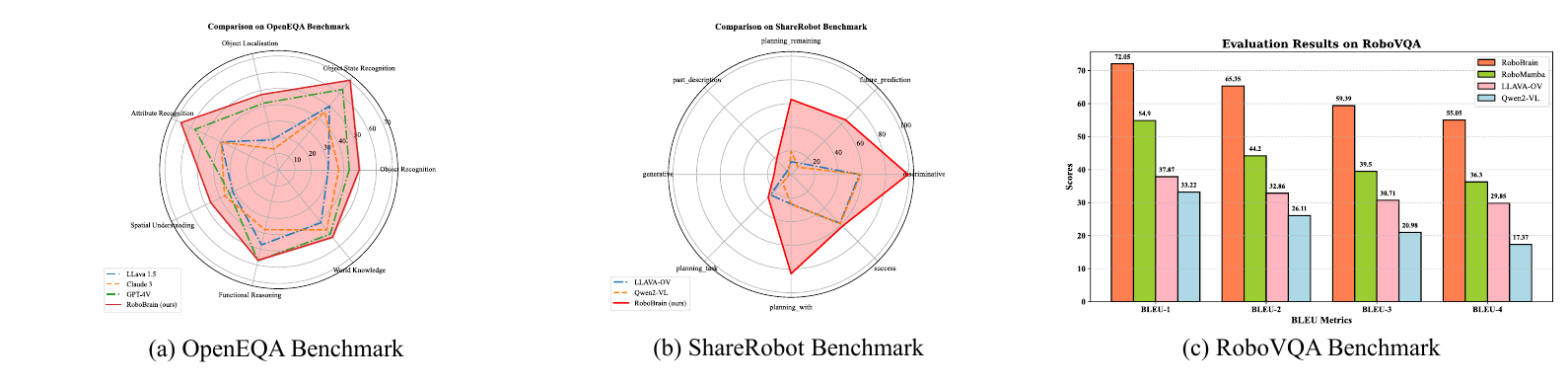}
    \caption{The performance of our model RoboBrain on the OpenEQA, ShareRobot, and RoboVQA benchmarks.  RoboBrain surpassed all baseline models, achieving state-of-the-art results.}
    \label{fig:radar}
\vspace{-0.5em}
\end{figure*}

\textbf{Phase 2: Robotic Training}
In Phase 2, we build upon the robust multi-modal foundational model developed in Phase 1 to create a more powerful model for robotic manipulation planning. Specifically, we aim for RoboBrain to understand complex, abstract instructions, support the perception of historical frame information and high-resolution images, and output object affordance regions while predicting potential manipulation trajectories. This will facilitate the transition from abstract to concrete in manipulation planning tasks. Detailed information is provided in Tab.~\ref{tab:training_strategy}.

In Stage 3, we collected a dataset of 1.3M robotic data to improve the model's manipulation planning capabilities. Specifically, this data is sourced from RoboVQA-800K~\cite{RoboVQA}, \textit{ScanView-318K} including MMScan-224K~\cite{mmscan,leo}, 3RScan-43K~\cite{3rscan,leo}, ScanQA-25K~\cite{scanqa,leo}, SQA3d-26K~\cite{sqa3d,leo}, and a subset of ShareRobot-200K introduced in this paper. These datasets contain substantial scene-scanning image data, long video data, and high-resolution data to support the model's ability to perceive diverse environments. Additionally, the fine-grained, high-quality planning data in the ShareRobot dataset enhances the manipulation planning capabilities of RoboBrain. To mitigate the issue of catastrophic forgetting~\cite{forget}, we selected a high-quality subset of approximately 1.7M image-text data from Phase 1 to mix with the robotic data collected in Stage 3 for training, tuning the entire model accordingly. In Stage 4, we enhanced the model's ability to perceive object affordances and predict manipulation trajectories from instructions, utilizing affordance and trajectory data from the ShareRobot dataset and other open-source sources~\cite{AGD20K,niu2024llarva}. This was achieved by incorporating LoRA modules during training for concrete manipulation capabilities.

%% file: sec/4_experiment.tex
\begin{figure*}[!t]
    \centering
    \includegraphics[width=0.95\linewidth]{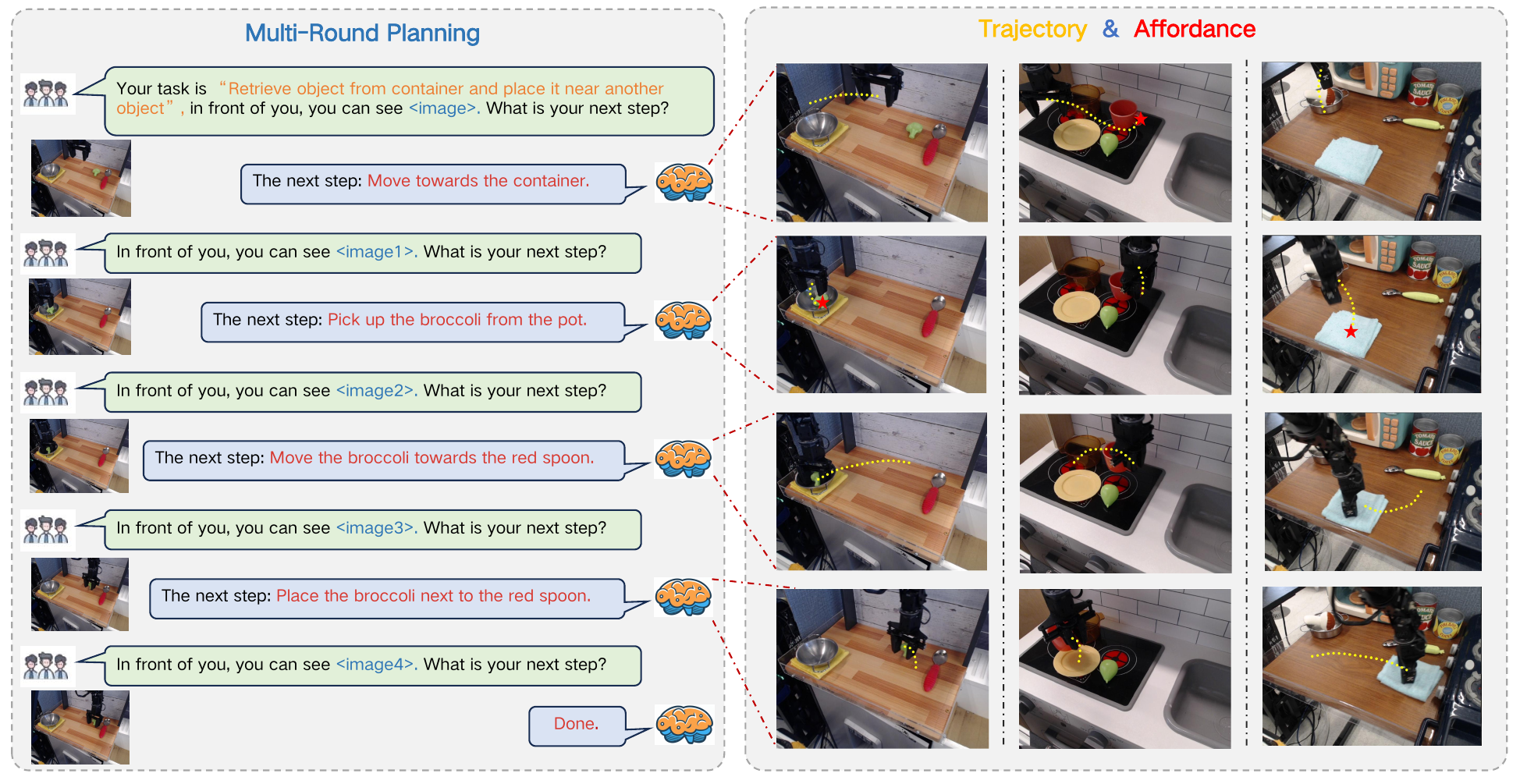}
    \caption{    
    This visualization illustrates that RoboBrain can interpret human instructions and visual images to generate action plans and assessments based on real-time image feedback. Furthermore, it predicts trajectories for each step and identifies corresponding affordances.}
    \label{fig:visualization}
\end{figure*}

\section{Experiment}
\subsection{Implementation Details}

During the entire training phase, we employed the Zero3~\cite{zero} distributed training strategy, conducting all experiments on a cluster of servers, each equipped with 8$\times$A800 GPUs. The training components for each stage, including image resolution settings, batch size, epochs, and learning rates, are provided in Tab.~\ref{tab:training_strategy}.

\subsection{Evaluation Metrics}
\textbf{Planning Task}
We selected RoboVQA~\cite{RoboVQA}, OpenEQA~\cite{OpenEQA}, and the test set of ShareRobot as robotic benchmarks for multi-dimensional assessment. For RoboVQA, we adopt the BLEU1 to BLEU4 metrics~\cite{bleu} used in RoboMamba~\cite{Robomamba} for evaluation. Additionally, for OpenEQA and ShareRobot, we use GPT-4o~\cite{gpt4o} as the evaluation tool, scoring based on the alignment or similarity between model predictions and ground truth, which serves as the final performance score for the model.

\textbf{Affordance Prediction} We utilize the Average Precision (AP) to evaluate the affordance performance of our model. AP metric summarizes the precision-recall affordance curve, which plots the relationship between precision and recall at various threshold settings. It is calculated across multiple Intersection over Union (IoU) thresholds to obtain a more comprehensive evaluation.

\textbf{Trajectory Prediction} We evaluate the similarity between ground truth and predicted trajectories, both represented as sequences of 2D waypoints normalized to $[0, 1000)$, following Qwen2-VL \cite{wang2024qwen2}. The evaluation uses three metrics: Discrete Fréchet Distance (DFD) \cite{gu2023rt}, Hausdorff Distance (HD), and Root Mean Square Error (RMSE). DFD captures overall shape and temporal alignment, HD identifies maximum deviation, and RMSE measures average pointwise error. Together, these metrics provide a comprehensive assessment of trajectory accuracy and similarity.

\subsection{Evaluation on Robot Brain Task}
\textbf{Evaluation on Planning Task} 
We selected 6 powerful MLLMs as our baselines for comparison, including both open-source and closed-source models with different architectures. Specifically, these models include GPT-4V~\cite{gpt4v}, Claude3~\cite{Claude}, LLaVA-1.5~\cite{LLaVa}, LLaVA-OneVision-7b~\cite{li2024llava}, Qwen2-VL-7b~\cite{qwen2vl} and RoboMamba~\cite{Robomamba}.
Our specific experimental results are shown in Fig.~\ref{fig:radar}. Our RoboBrain outperformed all baseline models across three robotic benchmarks. RoboBrain significantly outperformed all baseline models on OpenEQA and ShareRobot, which can be attributed to its robust capabilities in understanding robotic tasks and perceiving long videos. Additionally, this pattern was observed in other benchmarks as well, with RoboBrain consistently demonstrating superior performance on RoboVQA, achieving a BLEU-4 score that exceeded that of the second-place model by 18.75. This result highlights its capability to decompose complex long-range task planning.

\textbf{Evaluation on Affordance Prediction } Our results are summarized in Tab.~\ref{tab:affordance}. We compare the Qwen2-VL-7B and LLaVA-NeXT-7B models. Qwen2-VL \cite{qwen2vl} has a superior visual grounding ability and LLaVA-NeXT \cite{llavanext} owns a high-resolution and strong vision tower. We test them all on the AGD20K affordance test set. Our RoboBrain outperforms significantly the other models. It surpasses Qwen2-VL \cite{qwen2vl} by $14.6$ AP, and LLaVA-NeXT by $17.3$ AP. It validates our RoboBrain can understand the physical properties of objects and provide the affordance accurately.

\begin{table}[!t]
    \centering
    \caption{\textbf{The comparison of affordance prediction.} We utilize AP as the metric, and test them on affordance test set.}
    \label{tab:affordance}
    \begin{tabular}{l|c}
        \hline
        \textbf{Model} & \textbf{AP} $\uparrow$  \\ 
        \hline
        LLaVA-NeXT-7B \cite{LLaVa}  & 9.8 \%  \\
        Qwen2-VL-7B \cite{Qwen}  & 12.5 \% \\
        \rowcolor{blue!10} 
        RoboBrain (Ours) & \textbf{27.1} \% (14.6$\uparrow$) \\
        \hline
    \end{tabular}
    \vspace{-1em}
\end{table}

\textbf{Evaluation on Trajectory Prediction} 
We compare several variants of our model, and the results are in Tab.~\ref{table:evaluation_trajectory}: \textbf{(1) Baseline}, fine-tuned on trajectory-related VQA data; \textbf{(2) Start\_Points}, which adds the 2D start coordinates of the end-effector; \textbf{(3) Max\_Points}, limiting waypoints to 10 via uniform sampling; and \textbf{(4) Spec\_Token \& End\_Points}, which adds end-effector positions and special tokens to emphasize waypoints and start/goal points. Each variant builds on the previous one, with the final model integrating all components. Our most effective model integrates all design choices. As shown in the last row of Tab.~\ref{table:evaluation_trajectory}, DFD, HD, and RMSE decreased by 42.9\%, 94.2\%, and 31.6\%, respectively, compared to the baseline. We found that adding start points corrected the translational offset between the generated trajectory and the end-effector.

\begin{table}[t]
    \centering
    \renewcommand{\arraystretch}{1.2}
    \setlength{\tabcolsep}{3pt}
    \caption{\textbf{Trajectory Prediction Results Comparison}. Discrete Fréchet Distance (DFD), Hausdorff Distance (HD), and Root Mean Square Error (RMSE).}
    \label{table:evaluation_trajectory}
    \resizebox{0.40\textwidth}{!}{%
    \begin{tabular}{lccc}
        \toprule
        \textbf{Method} & \textbf{DFD $\downarrow$} & \textbf{HD $\downarrow$} & \textbf{RMSE $\downarrow$} \\
        \midrule
        RoboBrain (Base) & 0.191 & 0.171 & 0.133 \\
        \hspace{1em}+ Start\_Points & 0.176 & 0.157 & 0.117 \\
        \hspace{1em}+ Max\_Points & 0.185 & 0.163 & 0.125 \\
        \rowcolor{blue!10}
        \hspace{1em}+ Spec\_Token & \textbf{0.109} (42.9\%$\downarrow$) & \textbf{0.010} (94.2\%$\downarrow$) & \textbf{0.091} (31.6\%$\downarrow$) \\
        \bottomrule
    \end{tabular}%
    }
    \vspace{-1em}
\end{table}

\subsection{Visualization}
In this section, we present visual examples of RoboBrain in Fig.~\ref{fig:visualization}. Given human instructions and visual inputs, RoboBrain engages in multi-turn interactions, understanding and planning future steps. It also outputs more concrete affordances and trajectories.

%% file: sec/6_conclusion.tex
\vspace{-0.5em}
\section{Conclusion}
In this paper, we introduce \textbf{\textit{ShareRobot}}, a high-quality dataset that labels multi-dimensional information, including task planning, object affordance, and end-effector trajectory. We also present \textbf{\textit{RoboBrain}}, an MLLM-based model that integrates robotic and general multi-modal data, employs a multi-stage training strategy, and leverages long videos and high-resolution images to enhance robotic manipulation. Extensive experiments demonstrate that RoboBrain achieves state-of-the-art performance across various robotic tasks, underscoring its potential to significantly advance robotic capabilities.

\section*{Acknowledgments}
{\raggedright
This work was supported by the National Natural Science Foundation of China 62476011, 72225011 and 72434005.
\par}

%% file: sec/X_suppl.tex
\newpage
\appendix
\section*{Appendix}
This supplementary material provides more details of the proposed method and experiment results that are omitted from the manuscript due to the page limit. 
Sec.~\ref{sec1} presents additional details of the models and training strategies. 
Sec.~\ref{sec2} presents details of training dataset.
Sec.~\ref{sec3} complements more experiment results and analysis. 
Sec.~\ref{sec4} shows more visualization results to
prove the effectiveness of RoboBrain.
Sec.~\ref{sec-data} introduces more details about the construction of ShareRobot dataset.
Sec.~\ref{future} discusses potential future research directions for RoboBrain.

\section{Details of Models and Training}
\label{sec1}
\textbf{Model Setting.}
RoboBrain is built upon the LLaVA~\cite{LLaVa} framework and consists of three main components: the visual encoder, projector, and large language model (LLM).

For the visual encoder, we utilized the SigLIP~\cite{siglip} model, specifically the siglip-so400m-patch14-384, which is pre-trained on WebLi~\cite{pali} at a resolution of 384x384. The SigLIP model improves upon traditional CLIP~\cite{clip,hao2023mixgen} architectures by employing a sigmoid loss function that operates solely on image-text pairs, eliminating the need for global normalization of pairwise similarities. This enhancement allows for more efficient scaling of batch sizes while maintaining performance, even at smaller scales. SigLIP has 27 hidden layers and processes input images using patches of size 14x14, resulting in 729 visual tokens per image. The projector consists of a two-layer MLP~\cite{llavanext} that projects the visual tokens obtained from the visual encoder to the dimensions of the text embeddings. For the LLM, we adopted the Qwen2.5-7B-Instruct~\cite{qwen2_5} model, which is a state-of-the-art open-source LLM that is part of the latest Qwen series~\cite{Qwen}. It features 28 hidden layers and supports long-context inputs of up to 128K tokens, providing multilingual capabilities across more than 29 languages. 

In Stage 4, we introduced LoRA~\cite{lora} to train RoboBrain, enabling it to acquire affordance perception and trajectory prediction capabilities. LoRA allows for parameter-efficient fine-tuning of large models by adding low-rank parameter matrices to existing layers. We incorporated LoRA modules with a rank of 64 into the feed-forward network layers of both the Projector and the LLM, freezing all parameters except those of the LoRA modules during training.

\textbf{Training Setting.}
In the main text of the paper, we employed a staged training strategy, with complete settings presented in Tab.~\ref{tab:training_setting_appendix}. We primarily referenced the training strategy of LLaVA-Onevision~\cite{li2024llava}, a state-of-the-art multimodal large language model, and built upon this foundation to expand the robotic training phase. During the entire training phase, we conducted all experiments on a cluster of servers, each equipped with 8$\times$A800 GPUs.

\begin{table*}[t]
    \centering
    \caption{Detailed configuration for each training stage of the RoboBrain.}
    \label{tab:training_setting_appendix}
    \setlength{\tabcolsep}{12pt}
    \renewcommand{\arraystretch}{1.2}
    \resizebox{\textwidth}{!}{%
    \begin{tabular}{@{}ll|c|c|c|c|c|c|c@{}}
        \toprule
        & & \textbf{Stage-1} & \textbf{Stage-1.5} & \multicolumn{2}{c|}{\textbf{Stage-2}} & \textbf{Stage-3} & \multicolumn{2}{c}{\textbf{Stage-4}} \\ \cmidrule(l){5-6} \cmidrule(l){8-9}
        & & & & \textbf{Single-Image} & \textbf{OneVision} &  & \textbf{A-LoRA} & \textbf{T-LoRA} \\
        \midrule 
        \multirow{2}{*}{\rotatebox[origin=c]{90}{\small \textit{Vision}}}
        & \textbf{Resolution}  & 384 & Max 384\footnotesize{$\times$\{2$\times$2\}} & Max 384\footnotesize{$\times$\{6$\times$6\}} & Max 384\footnotesize{$\times$\{6$\times$6\}} & Max 384\footnotesize{$\times$\{6$\times$6\}} & Max 384\footnotesize{$\times$\{6$\times$6\}} & Max 384\footnotesize{$\times$\{6$\times$6\}} \\
        & \#Tokens & 729 & Max 729\footnotesize{$\times$5} & Max 729\footnotesize{$\times$37} & Max 729{\footnotesize{$\times$37}} & Max 729{\footnotesize{$\times$37}} & Max 729{\footnotesize{$\times$37}} & Max 729{\footnotesize{$\times$37}} \\
        \midrule 
        \multirow{2}{*}{\rotatebox[origin=c]{90}{\small \textit{Model}}}
        & \textbf{Trainable} & Projector & Full Model & Full Model & Full Model & Full Model & A-LoRA & T-LoRA \\
        & \#Tunable Parameters & 17.0M & 8.0B & 8.0B & 8.0B & 8.0B & 28.0M & 28.0M \\
        \midrule 
        \multirow{14}{*}{\rotatebox[origin=c]{90}{\small \textit{Training}}}
        & \textbf{Per-device Batch Size} & 8 & 2 & 1 & 1 & 1 & 4 & 4 \\
        & \textbf{Gradient Accumulation} & 1 & 2 & 2 & 2 & 2 & 2 & 2 \\
        & \textbf{LR: $\psiv_{\text{ViT}}$} & - & 2 $\times 10^{-6}$ & 2 $\times 10^{-6}$ & 2 $\times 10^{-6}$ & 2 $\times 10^{-6}$ & 2 $\times 10^{-6}$ & 2 $\times 10^{-6}$ \\    
        & \textbf{LR: $\{\thetav_{\text{Proj.}}, \phiv_{\text{LLM}}, \phiv_{\text{LoRA}}\}$} & 1$\times 10^{-3}$ & 1 $\times 10^{-5}$ & 1 $\times 10^{-5}$ & 1 $\times 10^{-5}$ & 1 $\times 10^{-5}$ & 1 $\times 10^{-5}$ & 1 $\times 10^{-5}$ \\
        & \textbf{Epoch} & 1 & 1 & 1 & 1 & 1 & 1 & 1 \\
        & \textbf{Optimizer} & AdamW & AdamW & AdamW & AdamW & AdamW & AdamW & AdamW \\
        & \textbf{Deepspeed} & Zero3 & Zero3 & Zero3 & Zero3 & Zero3 & Zero2 & Zero2 \\
        & \textbf{Weight Decay} & 0 & 0 & 0 & 0 & 0 & 0 & 0 \\
        & \textbf{Warmup Ratio} & 0.03 & 0.03 & 0.03 & 0.03 & 0.03 & 0.03 & 0.03 \\
        & \textbf{LR Schedule} & cosine & cosine & cosine & cosine & cosine & cosine & cosine \\
        & \textbf{Projector Type} & mlp2x\_gelu & mlp2x\_gelu & mlp2x\_gelu & mlp2x\_gelu & mlp2x\_gelu & mlp2x\_gelu & mlp2x\_gelu \\
        & \textbf{Vision Select Layer} & -2 & -2 & -2 & -2 & -2 & -2 & -2 \\
        & \textbf{Patch Merge Type} & spatial\_unpad & spatial\_unpad & spatial\_unpad & spatial\_unpad & spatial\_unpad & spatial\_unpad & spatial\_unpad \\
        & \textbf{Frames Upbound} & - & - & - & 32 & 32 & 32 & 32 \\
        & \textbf{Max Seq Length} & 8192 & 32768 & 32768 & 32768 & 32768 & 4096 & 4096 \\
        & \textbf{GPU Nums} & 16*8 & 16*8 & 20*8 & 20*8 & 22*8 & 4*8 & 4*8 \\
        \bottomrule
    \end{tabular}
    }
    \vspace{1mm}
    \vspace{-0.5em}
\end{table*}

\section{Details of Training Dataset}
\label{sec2}
\begin{figure}[t]
    \centering
    \includegraphics[width=0.92\linewidth]{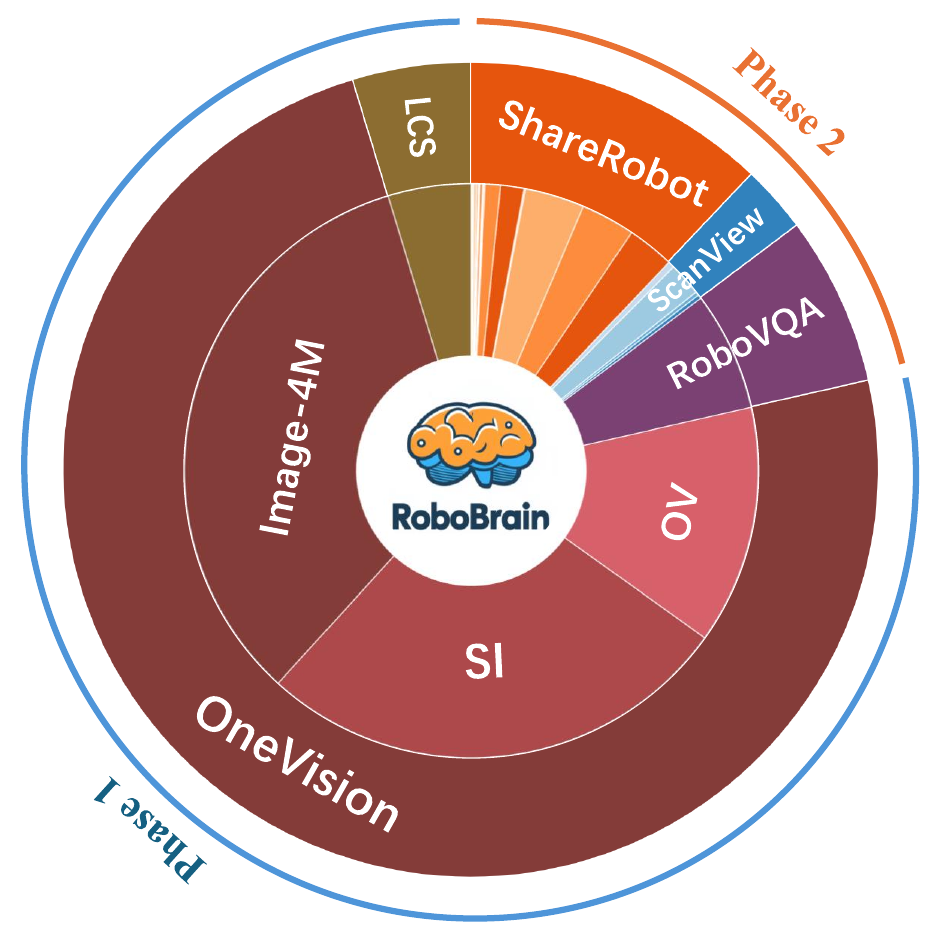}
    \caption{The distribution of the entire training dataset.}
    \label{fig:training_data}
\end{figure}

In the main body of the paper, we emphasize the importance of the training data and the proportion of robotic data. In this section, we will provide a detailed overview of the training data and its sources. The distribution of the entire training dataset is illustrated in Fig.~\ref{fig:training_data}.

\begin{itemize}
    \item
    \textbf{LCS-558K} is a subset of the LAION/CC/SBU dataset \cite{cc3m,laion}, specifically designed as the LLaVA Visual Instruct Pretrain~\cite{LLaVa} Dataset. This dataset has been filtered to ensure a balanced distribution of concept coverage, providing diverse and representative visual content. The primary purpose of LCS-558K is to facilitate alignment between the visual encoder and the LLM, enabling the LLM to understand visual information.
    \item
    \textbf{Image-4M} comprises 8 data sources, including 3 from the LLaVA-Recap series~\cite{li2024llava-next}: BLIP558K, COCO118K, and CC3M, as well as UReader~\cite{ye2023ureader}, Instruct Azure DC~\cite{li2024llava-next}, Evol-Instruct~\cite{chen2024allava}, and SynthDog~\cite{kim2022donut} We utilized the download links provided by the LLaVA-OneVision team for the data acquisition.
    \item
    \textbf{SI-3.2M}\footnote{Due to the unavailability of certain datasets, the actual data used amounts to 3.1M.}~\cite{li2024llava} consists of 3.2 million samples, carefully curated to support multimodal learning. It includes subsets from existing datasets such as Cambrian~\cite{tong2024cambrian}, Cauldron~\cite{li2024llava-next}, and UReader~\cite{ye2023ureader}, which were subjected to cleaning and re-annotation to ensure data quality. Additionally, it incorporates single-image data from sources like AI2D~\cite{ai2d} and OKVQA~\cite{marino2019okvqa}, alongside a newly compiled single-image collection designed to achieve a balanced and diverse dataset.
    \item
    \textbf{OV-1.6M}\footnote{Due to the vague descriptions and missing key information regarding dataset filtering in the original paper, we ended up using 2.4M data.}~\cite{li2024llava} comprises 1.6 million samples, which includes approximately 800K high-quality samples re-sampled from earlier SI-3.2M datasets with a data replay strategy, ensuring improved data reliability and relevance. Additionally, the dataset incorporates M4-Instruct data to enrich instructional learning tasks. A significant component of OV-1.6M is its video data, which has been released alongside LLaVA-video data. The video subset used in the dataset is specifically aligned with the previous annotation format, providing a diverse multimodal resource for advancing vision-language learning.
    \item
    \textbf{RoboVQA-800K}~\cite{RoboVQA} consists of realistic data gathered from various user requests, utilizing different embodiments including robots, humans, and humans equipped with grasping tools. The dataset features 5,246 long-horizon episodes and 92,948 medium-horizon episodes of robotic tasks, with each episode accompanied by corresponding image and text prompt inputs. The primary purpose of RoboVQA-800K is to enhance RoboBrain's reasoning capabilities in robotic-related scenarios.
    \item
    \textbf{ScanView-318K} totals 318K samples, which integrates data from several high-quality sources, including MMScan-224K~\cite{mmscan}, 3RScan-43K~\cite{3rscan}, ScanQA-25K~\cite{scanqa}, and SQA3D-26K~\cite{sqa3d}, each contributing unique strengths. MMScan-224K provides multimodal scene data with detailed annotations, such as object segmentation and textual descriptions. 3RScan-43K offers 3D reconstructions and semantic annotation. ScanQA-25K includes question-answer pairs based on 3D scanned environments. SQA3D-26K focuses on spatial question answering. Together, these datasets provide diverse scene-scanning image data, long video sequences, and high-resolution samples, equipping models with fine-grained environmental perception and reasoning abilities.
\end{itemize}

\section{Complementary Experiments}
\label{sec3}
In this section, we present the complete experiments and results that are omitted from the manuscript due to page limitations. This includes an exploration of the impact of incorporating ShareRobot on training, the effects of varying proportions of robotic data in the training dataset, and more comprehensive results comparing RoboBrain with the baselines on both general and robotic benchmarks.

Additionally, we explore the impact of different architectures and pre-trained MLLMs, as well as different LLM backbones on our experimental results. We also conduct ablation studies at various stages to meticulously analyze the contributions of each stage to overall performance.

\begin{table*}[ht]
\centering
\renewcommand{\arraystretch}{1.25}
\caption{Performance comparison on multiple general benchmarks.}
\label{tab:suppl_general_benchmark}
\resizebox{\textwidth}{!} {
\begin{tabular}{llcccccc}
\toprule
\textbf{Dataset} & \textbf{Split} & \textbf{RoboBrain (Ours)} & \textbf{GPT-4V~\cite{gpt4v}} & \textbf{LLaVA-OV-7B~\cite{llavanext}} & \textbf{InternVL2-8B~\cite{internvl}} & \textbf{Qwen2-VL-7B~\cite{qwen2vl}} & \textbf{GPT-4o~\cite{gpt4o}} \\
\midrule
A12D\cite{ai2d}           & test      & 82.03       & 78.2    & 81.4           & 83.8           & -      & \textbf{94.2}     \\
ChartQA\cite{chartqa}        & test      & 80.48       & 78.5    & 80             & 83.3           & 83     & \textbf{85.7}     \\
DocVQA\cite{docvqa}         & test      & 88          & 88.4    & 87.5           & 91.6           & \textbf{94.5}   & 92.8     \\
TextVQA\cite{textvqa}        & val       & 75.85       & -       & 71.07          & 77.4           & \textbf{84.3}   & -        \\
MMMU\cite{mmmu}           & val       & 49          & 56.8    & 48.8           & 51.8           & 54.1   & \textbf{69.1}     \\
MMStar\cite{mmstar}         & test      & 61.23       & 57.1    & 61.7           & 61.5           & 60.7   & \textbf{63.9}     \\
OCRBench\cite{ocrbench}       & -         & 677         & 656     & 697            & 794            & \textbf{845}    & 805      \\
RealWorldQA\cite{realworldqa}    & test      & 68.89       & 61.4    & 66.3           & 64.4           & \textbf{70.1}   & 58.6     \\
SeedBench\cite{seedbench}      & image     & 71.03       & 49.9    & 75.4           & \textbf{76.2}           & -      & \textbf{76.2}     \\
MMbench\cite{mmbench}        & en-dev    & 81.52       & 81.3    & 83.2           & -              & -      & \textbf{83.4}     \\
MMbench\cite{mmbench}        & en-test   & 80.44       & 75      & 80.8           & 81.7           & \textbf{83}     & -        \\
MME\cite{fu2023mme}            & test      & 2084        & 1926    & 1998           & 2210           & \textbf{2327}   & -        \\
\bottomrule
\end{tabular}}
\vspace{-0.5em}
\end{table*}

\subsection{More Results on General Benchmarks}
\label{types}
To evaluate performance on general tasks in real-world scenarios, as is commonly done with MLLMs~\cite{gpt4v, llavanext, internvl, qwen2vl, gpt4o}, we conducted experiments using a diverse set of image benchmarks summarized in Table~\ref{tab:suppl_general_benchmark}. We leveraged the comprehensive evaluation toolkit, LMMs-Eval\cite{lmms, zhang2024unveiling}, to evaluate RoboBrain's performance on general benchmarks. These benchmarks are categorized into three classes:

\begin{itemize}
\item
\textbf{Chart, Diagram, and Document Understanding.}  
As key visual formats for structured OCR data, benchmarks such as AI2D~\cite{ai2d}, ChartQA~\cite{chartqa}, DocVQA~\cite{docvqa}, and OCRBench~\cite{ocrbench} were utilized. Open-source models like InternVL2-8B~\cite{internvl} and LLAVA-OV-7B~\cite{llavanext} have demonstrated comparable performance to closed-source models such as GPT-4V~\cite{gpt4v}. For \textit{RoboBrain}, despite being optimized primarily for multidimensional robotic tasks, it surpasses LLAVA-OV-7B~\cite{llavanext} and GPT-4V~\cite{gpt4v} on these benchmarks, achieving a significant improvement in structured OCR tasks, with the only exceptions being DocVQA~\cite{docvqa}, where it performs slightly lower than GPT-4V~\cite{gpt4v}, and OCRBench~\cite{ocrbench}, where it falls slightly behind LLAVA-OV-7B~\cite{llavanext}.

\item
\textbf{Visual Perception and Multi-domain Reasoning.}  
This category focuses on complex visual perception and multidisciplinary reasoning tasks. Benchmarks for visual perception include MMStar~\cite{mmstar}, MMBench~\cite{mmbench}, and MME~\cite{fu2023mme}, while reasoning benchmarks include MMMU~\cite{mmmu} and SeedBench~\cite{seedbench}. \textit{RoboBrain} demonstrates comparable performance to GPT-4V~\cite{gpt4v} and LLAVA-OV-7B~\cite{llavanext} across multiple benchmarks.

\item
\textbf{Real-world Understanding and Interaction.}  
Evaluating MLLMs~\cite{gpt4v, llavanext, internvl, qwen2vl, gpt4o} as general-purpose assistants in real-world settings is crucial, as these scenarios extend beyond controlled environments. For this, the RealworldQA~\cite{realworldqa} benchmark was utilized. Results indicate that \textit{RoboBrain} not only outperforms open-source models like LLAVA-OV-7B~\cite{llavanext} and InternVL2-8B~\cite{internvl}, but also exceeds closed-source models such as GPT-4V~\cite{gpt4v} and GPT-4o~\cite{gpt4o}, showcasing its extensive knowledge base and strong generalization capabilities.
\end{itemize}

\begin{table*}[ht]
\renewcommand{\arraystretch}{1.35}
\centering
\caption{Performance comparison on RoboVQA, OpenEQA and ShareRobot Benchmarks.}
\label{tab:suppl_robotic_benchmark}
\resizebox{\textwidth}{!} {
\begin{tabular}{llccccc}
\toprule
\textbf{Dataset}   & \textbf{Split / Metric}            & \textbf{RoboBrain (Ours)} & \textbf{GPT-4V~\cite{gpt4v}} & \textbf{LLaVA-OV-7B~\cite{llavanext}} & \textbf{RoboMamba~\cite{Robomamba}} & \textbf{Qwen2-VL-7B~\cite{qwen2vl}} \\
\midrule
\multirow{4}{*}{RoboVQA\cite{RoboVQA}}   & BLEU1                      & \textbf{72.05}  &   32.23   & 38.12   & 54.9  & 33.22       \\
                           & BLEU2                      & \textbf{65.35}  &   26.51   & 33.56   & 44.2  & 26.11       \\
                           & BLEU3                      & \textbf{59.39}  &   24.65   & 31.76   & 39.5  & 20.98       \\
                           & BLEU4                      & \textbf{55.05}  &   23.94   & 30.97   & 36.3  & 17.37       \\
\midrule
\multirow{7}{*}{OpenEQA\cite{OpenEQA}}   & OBJECT-STATE-RECOGNITION   & 70.4   & 63.2  & 72.02   & -     & \textbf{72.06}       \\
                           & OBJECT-RECOGNITION         & 49.54  & 43.4  & 51.73   & -     & \textbf{61.91}       \\
                           & FUNCTIONAL-REASONING       & 57.14  & \textbf{57.4}  & 55.53   & -     & 54.23       \\
                           & SPATIAL-UNDERSTANDING      & 46.46  & 33.6  & 48.98   & -     & \textbf{50.39}       \\
                           & ATTRIBUTE-RECOGNITION      & 66.7   & 57.2  & \textbf{75.52}   & -     & 73.88       \\
                           & WORLD-KNOWLEDGE            & 53.12  & 50.7  & 56.46   & -     & \textbf{57.3}        \\
                           & OBJECT-LOCALIZATION        & \textbf{47.45}  & 42    & 45.25   & -     & 47.29       \\
\midrule
\multirow{7}{*}{ShareRobot (Eval)} & DISCRIMINATIVE           & \textbf{99.02} &   -   & 57.9    & -     & 76.47          \\
                           & FUTURE-PREDICTION         & \textbf{72.92} &   -   & 13.1    & -     & 8.04           \\
                           & GENERATIVE                & \textbf{32.43} &   -   & 5.44    & -     & 4.63           \\
                           & PAST-DESCRIPTION          & \textbf{37.07} &   -   & 4.4     & -     & 13.65          \\
                           & PLANNING-REMAINING        & \textbf{71.29} &   -   & 24.5    & -     & 7.56           \\
                           & PLANNING-TASK             & \textbf{52.43} &   -   & 25      & -     & 36.34          \\
                           & PLANNING-WITH             & \textbf{91.95} &   -   & 44.25   & -     & 45.12          \\
                           & SUCCESS                   & \textbf{61.7}  &   -   & 58.5    & -     & 54.63          \\
\bottomrule
\end{tabular}}
\end{table*}

\subsection{More Results on Robotic Benchmarks.}

To evaluate \textit{RoboBrain}'s performance on robotic capabilities in real-world scenarios, we selected RoboVQA~\cite{RoboVQA}, OpenEQA~\cite{OpenEQA}, and the test set of ShareRobot, extracted from the proposed ShareRobot dataset, as robotic benchmarks for multi-dimensional assessment, as shown in Table~\ref{tab:suppl_robotic_benchmark}. The chosen baselines include MLLMs such as GPT-4V~\cite{gpt4v}, LLaVA-OV-7B~\cite{llavanext}, and Qwen2-VL-7B~\cite{qwen2vl}, as well as robotic models like RoboMamba~\cite{Robomamba}. Detailed descriptions of the three selected robotic benchmarks and the analysis of each results are provided below:

\begin{itemize}
\item 
\textbf{RoboVQA}~\cite{RoboVQA} provides a robotics VQA benchmark and a long-horizon planning benchmark with an intervention mechanism on real robots. Specifically, this benchmark includes 18,248 video-text pairs designed from 100 long-horizon episodes for various robotic VQA tasks, including planning, planning with context, planning remaining steps, future prediction, generative affordance, past description, success (positive/negative), and discriminative affordance (positive/negative). Similar to RoboMamba~\cite{Robomamba}, we utilized BLEU-1$\sim$BLEU-4 to evaluate the average performance across all tasks. According to the evaluation results, our proposed model, \textit{RoboBrain}, outperforms all baselines, achieving approximately 30\% higher performance than the second-best model.

\item 
\textbf{OpenEQA}~\cite{OpenEQA} provides a robotics VQA benchmark with over 1,600 high-quality human-generated questions drawn from more than 180 real-world scenes, targeting the task of Embodied Question Answering (EQA) for environment understanding. For fairness, we evaluated all models using the prompt templates and the LLM-Score metric provided by OpenEQA~\cite{OpenEQA}. Based on the evaluation results, our proposed model, \textit{RoboBrain}, outperforms GPT-4V~\cite{gpt4v} overall and achieves comparable performance to other baselines. In the future, we plan to further enhance \textit{RoboBrain}'s spatial intelligence to improve its generalization across scenes.
 
\item 
\textbf{ShareRobot (Eval)} provides a cross-scene and cross-embodiment robotics benchmark consisting of 2,050 VQA pairs, drawn from 102 diverse scenes (e.g., bedroom, laboratory, kitchen, office) and covering 12 different robot bodies. Similar to RoboVQA~\cite{RoboVQA}, we categorized various robotic VQA tasks into planning, planning with context, planning remaining steps, future prediction, generative affordance, past description, success (positive/negative), and discriminative affordance (positive/negative). Unlike RoboVQA benchmark~\cite{RoboVQA}, we utilized GPT-4o~\cite{gpt4o} to score the evaluation results instead of BLEU metrics for each task, aiming for more accurate performance assessment. Based on the results, our proposed model, \textit{RoboBrain}, outperforms all baselines, demonstrating its exceptional planning capabilities across diverse scenes and embodiments.

\end{itemize}

\subsection{Effectiveness of ShareRobot}
In this subsection, we investigate the effectiveness of the proposed ShareRobot dataset for training RoboBrain. We maintain the ratio of robotic data to general data used in the main body of the paper, approximately 4:6. Based on the original data source proportions, we randomly sampled 200K samples, which include:

\begin{itemize}
    \item \textbf{Exp A} consists of 40\% robotic data, with 20\% sourced from ShareRobot and 20\% from other robotic sources, along with 60\% general data.
    \item \textbf{Exp B} consists of 40\% robotic data, excluding ShareRobot, with the same other robotic data resampled as in Experiment A, resulting in a total of 40\%. It also includes 60\% general data, which is identical to that of Exp A.
\end{itemize}

We conducted a complete epoch for all the experiments mentioned above. The results are presented in Tab~\ref{tab:exp_data_distribution}. As shown in the table, the inclusion of ShareRobot data enhances the model's performance compared to scenarios without ShareRobot. This highlights ShareRobot's key role in enhancing RoboBrain's planning capabilities.

\begin{table*}[ht]
\renewcommand{\arraystretch}{1.35}
\centering
\caption{EExperimental results demonstrating the effectiveness of different task types. Type-1 refers to Chart, Diagram, and Document Understanding; Type-2 pertains to Visual Perception and Multi-domain Reasoning; Type-3 encompasses Real-world Understanding and Interaction. For detailed task descriptions, please refer to \ref{types}.}
\label{tab:exp_data_distribution}
\resizebox{\textwidth}{!} {
\begin{tabular}{c|c|cc|ccc|ccc|c}
\toprule
\multirow{2}{*}{\textbf{Exp. Name}} & \textbf{General Data (\%)} & \multicolumn{2}{c|}{\textbf{Robotic Data (\%)}} & \multicolumn{3}{c|}{\textbf{General Benchmarks}} & \multicolumn{3}{c|}{\textbf{Robotic Benchmarks}} & \multirow{2}{*}{\textbf{Average}}\\ 
\cmidrule{2-2} \cmidrule(lr){3-4} \cmidrule(lr){5-7} \cmidrule{8-10} 
                   & \textbf{OneVision} & \textbf{ShareRobot} & \textbf{Others} & \textbf{Type-1} & \textbf{Type-2} & \textbf{Type-3} & \textbf{RoboVQA}\cite{RoboVQA}& \textbf{OpenEQA}\cite{OpenEQA}& \textbf{ShareRobot}\\ 
\midrule
\textbf{EXP A} & 60\% & 20\% & 20\% & 62.44 & 71.98 & 70.33 & 48.29 & 58.74 & 63.11 & \textbf{\textcolor{blue}{62.48}} \\
\textbf{EXP B} & 60\% & 0\%  & 40\% & 62.36 & 71.38 & 66.01 & 49.20 & 57.96 & 27.03 & \textbf{55.66} \\
\midrule
\textbf{EXP C} & 70\% & 15\% & 15\% & 62.73 & 72.19 & 68.10 & 45.96 & 56.59 & 61.73 & \textbf{61.22} \\
\textbf{EXP D} & 60\% & 20\% & 20\% & 62.44 & 71.98 & 70.33 & 48.29 & 58.74 & 63.11 & \textbf{\textcolor{blue}{62.48}} \\
\textbf{EXP E} & 50\% & 25\% & 25\% & 62.28 & 71.25 & 66.54 & 49.34 & 58.76 & 63.35 & \textbf{61.92} \\
\textbf{EXP F} & 40\% & 30\% & 30\% & 62.39 & 71.61 & 68.37 & 49.22 & 56.24 & 64.57 & \textbf{62.07} \\
\textbf{EXP G} & 30\% & 35\% & 35\% & 62.69 & 71.92 & 69.54 & 47.74 & 55.72 & 65.22 & \textbf{62.14} \\
\bottomrule
\end{tabular}}
\vspace{-0.5em}
\end{table*}

\vspace{-0.5em}
\subsection{Effectiveness of Robot Data Proportion}
In this subsection, we investigate the effectiveness of the ratio of robotic data (including ShareRobot) to general data used in training RoboBrain. We maintain a constant total training dataset size of 200K while varying the sampling proportions of robotic and general data. The configurations are as follows:

\begin{itemize}
    \item \textbf{Exp C} utilizes a ratio of 3:7, comprising 30\% robotic data and 70\% general data.
    \item \textbf{Exp D} utilizes a ratio of 4:6, comprising 40\% robotic data and 60\% general data, \textbf{same to Exp A}.
    \item \textbf{Exp E} utilizes a ratio of 5:5, with 50\% robotic data and 50\% general data.
    \item \textbf{Exp F} utilizes a ratio of 6:4, featuring 60\% robotic data and 40\% general data.
    \item \textbf{Exp G} utilizes a ratio of 7:3, containing 70\% robotic data and 30\% general data.
\end{itemize}

We conducted a complete epoch for all the experiments mentioned above. The results are presented in Tab~\ref{tab:exp_data_distribution}. As shown in the table, a 4:6 ratio of robotic data is an effective choice for training, balancing performance on both the robotic and general benchmarks.

\begin{table}[t]
\begin{center}
\caption{\textbf{Additional Experimental Results.} ``SFT Data (G:R)'' indicates the ratio of training data for fine-tuning MLLMs, where ``G'' represents general VQA data and ``R'' denotes robot data (with half being ShareRobot). The total dataset size is 1.47M.}
\footnotesize
\scalebox{0.7}{
  \begin{tabular}{p{0.1cm}p{2cm}p{1.8cm}|p{1.5cm}p{1.5cm}|p{0.7cm}p{1cm}}
  \hline
 \rowcolor{black!10} \makecell[c]{}&  \makecell[c]{Model}& \makecell[c]{SFT Data(G:R)} & \makecell[c]{RoboVQA} &\makecell[c]{ShareRobot}&\makecell[c]{MME}&\makecell[c]{MMMU}\\
  \midrule

    \multirow{2}{*}{\makecell{}} & \multirow{2}{*}{\makecell[c]{LLaVA-OV-7b}} 
     & \makecell[c]{6:0}& \makecell[c]{36.29} &\makecell[c]{27.04}&\makecell[c]{\textbf{2001}}&\makecell[c]{\textbf{49.65}}\\
    & & \cellcolor{blue!10}\makecell[c]{6:4}& \cellcolor{blue!10}\makecell[c]{\textbf{43.63}} &\cellcolor{blue!10}\makecell[c]{\textbf{54.66}}&\cellcolor{blue!10}\makecell[c]{1945}&\cellcolor{blue!10}\makecell[c]{48.83}\\

    \multirow{2}{*}{\makecell{(a)}} & \multirow{2}{*}{\makecell[c]{Qwen2VL-7B}} 
    & \makecell[c]{6:0}& \makecell[c]{24.05} &\makecell[c]{28.17}&\makecell[c]{\textbf{2313}}&\makecell[c]{52.10}\\
    & & \cellcolor{blue!10}\makecell[c]{6:4}& \cellcolor{blue!10}\makecell[c]{\textbf{58.94}} &\cellcolor{blue!10}\makecell[c]{\textbf{58.86}}&\cellcolor{blue!10}\makecell[c]{2295}&\cellcolor{blue!10}\makecell[c]{\textbf{52.33}}\\

    \multirow{2}{*}{\makecell{}} & \multirow{2}{*}{\makecell[c]{OpenVLA-7B}} 
    & \makecell[c]{6:0}& \makecell[c]{4.11} &\makecell[c]{21.44}&\makecell[c]{1681}&\makecell[c]{35.07}\\
    & & \cellcolor{blue!10}\makecell[c]{6:4}& \cellcolor{blue!10}\makecell[c]{\textbf{54.79}} &\cellcolor{blue!10}\makecell[c]{\textbf{60.56}}&\cellcolor{blue!10}\makecell[c]{\textbf{1722}}&\cellcolor{blue!10}\makecell[c]{\textbf{37.25}}\\
    \hline

    \multirow{2}{*}{\makecell{}} & \multirow{2}{*}{\makecell[c]{LLaVA1.5-Qwen}} 
    & \makecell[c]{6:0}& \makecell[c]{24.17} &\makecell[c]{26.73}&\makecell[c]{1720}&\makecell[c]{44.28}\\
    & & \cellcolor{blue!10}\makecell[c]{6:4}& \cellcolor{blue!10}\makecell[c]{\textbf{49.01}} &\cellcolor{blue!10}\makecell[c]{\textbf{43.41}}&\cellcolor{blue!10}\makecell[c]{\textbf{1732}}&\cellcolor{blue!10}\makecell[c]{\textbf{48.33}}\\

    \multirow{4}{*}{\makecell{(b)}} & \multirow{2}{*}{\makecell[c]{LLaVA1.5-LLaMA}} 
    & \makecell[c]{6:0}& \makecell[c]{21.40} &\makecell[c]{25.06}&\makecell[c]{1529}&\makecell[c]{46.40}\\
    & & \cellcolor{blue!10}\makecell[c]{6:4}& \cellcolor{blue!10}\makecell[c]{\textbf{49.67}} &\cellcolor{blue!10}\makecell[c]{\textbf{54.87}}&\cellcolor{blue!10}\makecell[c]{\textbf{1722}}&\cellcolor{blue!10}\makecell[c]{\textbf{43.41}}\\

    \multirow{2}{*}{\makecell{}} & \multirow{2}{*}{\makecell[c]{LLaVA1.5-Vicuna}} 
    & \makecell[c]{6:0}& \makecell[c]{26.19} &\makecell[c]{22.18}&\makecell[c]{\textbf{1668}}&\makecell[c]{30.09}\\
    & & \cellcolor{blue!10}\makecell[c]{6:4}& \cellcolor{blue!10}\makecell[c]{\textbf{50.40}} &\cellcolor{blue!10}\makecell[c]{\textbf{51.42}}&\cellcolor{blue!10}\makecell[c]{1650}&\cellcolor{blue!10}\makecell[c]{\textbf{31.51}}\\

    \multirow{2}{*}{\makecell{}} & \multirow{2}{*}{\makecell[c]{LLaVA1.5-Mistral}} 
    & \makecell[c]{6:0}& \makecell[c]{14.30} &\makecell[c]{21.88}&\makecell[c]{\textbf{1602}}&\makecell[c]{23.91}\\
    & & \cellcolor{blue!10}\makecell[c]{6:4}& \cellcolor{blue!10}\makecell[c]{\textbf{36.29}} &\cellcolor{blue!10}\makecell[c]{\textbf{57.47}}&\cellcolor{blue!10}\makecell[c]{1548}&\cellcolor{blue!10}\makecell[c]{\textbf{24.32}}\\

  \bottomrule
    \label{tab_additional}
  \end{tabular}}
\end{center}
\vspace{-1.5em}
\end{table}

\subsection{Different Architecture and MLLMs}
To validate the effectiveness of different architecture and pre-trained MLLMs and training data in the stage 3 training setup, we selected LLaVA-OV-7B~\cite{li2024llava}, OpenVLA-7B~\cite{OpenVLA}, and Qwen2VL-7B~\cite{qwen2vl}, each representing a distinct architecture among MLLMs, and conducted supervised fine-tuning (SFT) using the same proportion of training data described in the main text. As shown in Tab.~\ref{tab_additional}~(a), the results demonstrated that incorporating ShareRobot can significant performance improvements. For unaligned MLLMs such as LLaVA 1.5~\cite{llava-1.5} and OpenVLA, we first aligned the MLP using BLIP-558k~\cite{li2024llava-next} before fine-tuning. In contrast, other models that already have aligned vision encoder and LLM were directly fine-tuned.

\subsection{Different LLM Backbones}
To demonstrate the effectiveness of different LLM backbones when fine-tuned on the ShareRobot dataset, we conducted experiments using four distinct LLMs~\cite{Qwen, llama, vicuna, mistral}. These models were fine-tuned using the ShareRobot data, and the experimental results are summarized in Tab.~\ref{tab_additional}~(b). The findings indicate that different LLMs benefit from the ShareRobot data.

\subsection{Ablation Studies of Different Stages}
We present the evaluation results for each stage in Tab. \ref{eval}. The results demonstrate that staged training from stage 1 to stage 3 consistently and effectively improves the model's planning performance, while stage 4 enhances the model's affordance and trajectory capabilities.

\begin{table}[t]
\begin{center}
\caption{Additional Evaluation Results. ``S1.5'' refers to Stage 1.5, and this notation applies similarly to other stages.}
\footnotesize
\scalebox{0.63}{
  \begin{tabular}{p{1cm}p{1pt}|p{1.5cm}p{1.5cm}|p{1.5cm}p{1.5cm}p{1.5cm}p{1.5cm}}
  \hline

 \rowcolor{black!10}  \makecell[c]{Stage} && \makecell[c]{RoboVQA} &\makecell[c]{ShareRobot}&\makecell[c]{MME}&\makecell[c]{MMMU}&\makecell[c]{Affordance$\uparrow$}&\makecell[c]{Trajectory$\downarrow$}\\
  \midrule
    \makecell[c]{S1.5} && \makecell[c]{2.60} &\makecell[c]{9.81}&\makecell[c]{1406}&\makecell[c]{46.00}&\makecell[c]{0.00}&\makecell[c]{1.00}\\

    \makecell[c]{S2-si}&& \makecell[c]{28.90} &\makecell[c]{13.31}&\makecell[c]{\textbf{2110}}&\makecell[c]{\textbf{50.76}}&\makecell[c]{3.11}&\makecell[c]{1.00}\\

    \makecell[c]{S2-ov}&& \makecell[c]{31.81} &\makecell[c]{34.84}&\makecell[c]{2083}&\makecell[c]{49.95}&\makecell[c]{8.50}&\makecell[c]{1.00}\\

    \rowcolor{blue!10} \makecell[c]{S3}&& \makecell[c]{\textbf{62.96}} &\makecell[c]{\textbf{65.05}}&\makecell[c]{2084}&\makecell[c]{49.00}&\makecell[c]{7.14}&\makecell[c]{1.00}\\

    \rowcolor{blue!10} \makecell[c]{S4-A}&& \makecell[c]{\textbf{62.96}} &\makecell[c]{\textbf{65.05}}&\makecell[c]{2084}&\makecell[c]{49.00}&\makecell[c]{\textbf{27.1}}&\makecell[c]{-}\\

    \rowcolor{blue!10} \makecell[c]{S4-T}&& \makecell[c]{\textbf{62.96}} &\makecell[c]{\textbf{65.05}}&\makecell[c]{2084}&\makecell[c]{49.00}&\makecell[c]{-}&\makecell[c]{\textbf{0.09}}\\
  \bottomrule
  \label{eval}
  \end{tabular}}
\end{center}
\vspace{-3em}
\end{table}

\section{More Qualitative Results}
\label{sec4}
In this section, we provide additional visual results for planning, affordance perception, and trajectory prediction. This includes the presentation of both positive and negative samples, as well as further analysis.

\subsection{Visualization on Planning} 
\label{subsec4.1}
\begin{figure*}[!h]
    \centering
    \includegraphics[width=0.92\linewidth]{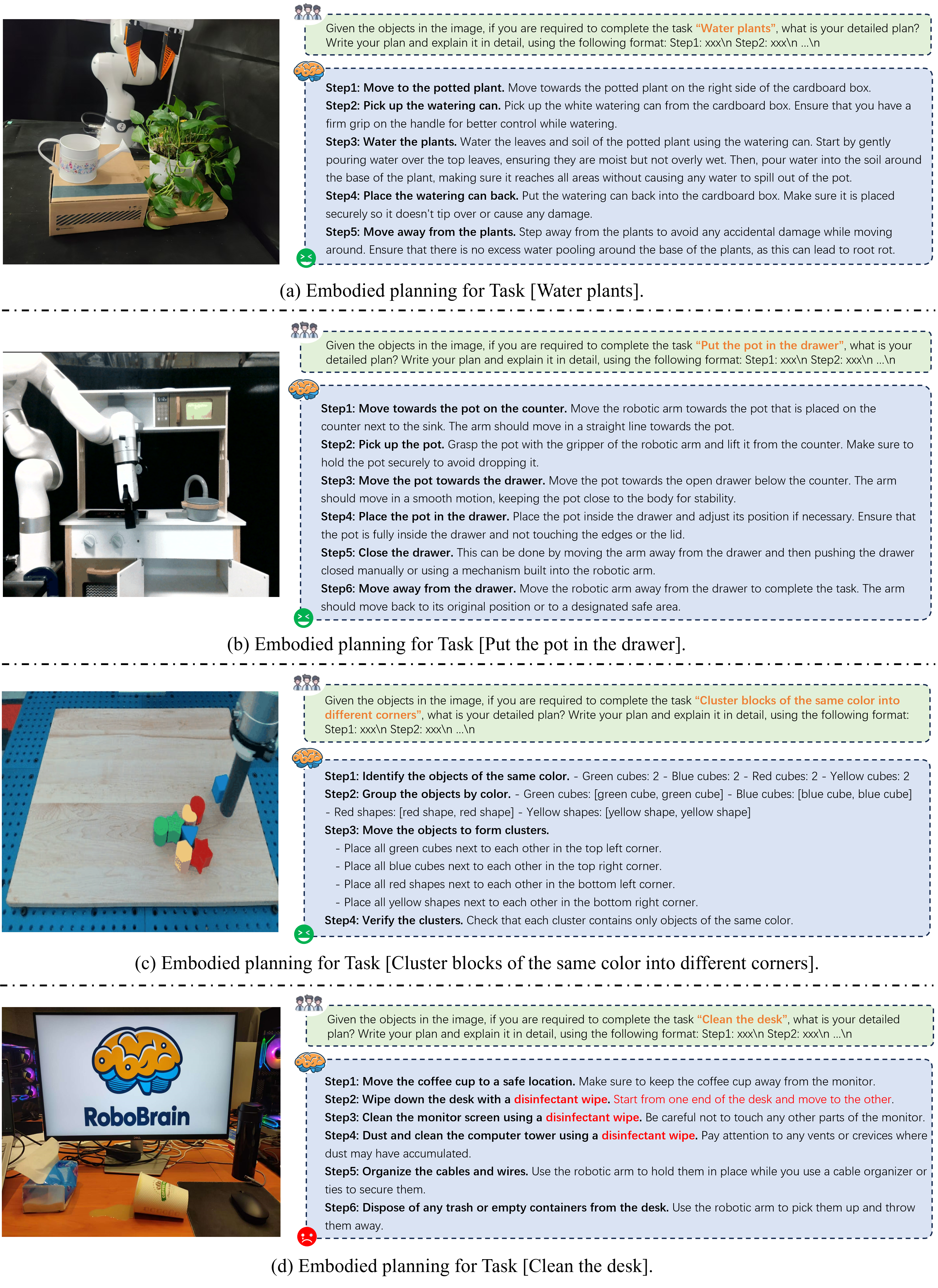}
    \caption{\textbf{Additional embodied planning of RoboBrain.} (a)$\sim$(c) show some good cases of RoboBrain's embodied planning, while (d) shows its bad case. More detailed analysis can be found in Sec.\ref{subsec4.1}.}
    \label{fig:suppl_planning_demo}
\end{figure*}

Here, we present additional embodied planning for robotic tasks generated by RoboBrain, as shown in Fig. \ref{fig:suppl_planning_demo}. In this figure, we demonstrate the planning results of RoboBrain for four distinct robotic manipulation tasks: "Water plants", "Put the pot in the drawer", "Cluster blocks of the same color into different corners", and "Clean the desk", where the first three are categorized as good cases, and the last one as a bad case. Additionally, the model provides a rationale and detailed explanation for each step of the planning process across all four cases.

From the first three planning cases, it is evident that RoboBrain effectively utilizes environmental information and the states of interactive objects—captured from first- or third-person perspective images—to generate task plans for various types of robotic manipulation tasks. Notably, in the "Cluster blocks of the same color into different corners" task, RoboBrain not only analyzes the number of blocks of each color on the table in Steps 1 and 2 but also provides detailed sub-steps in Step 3, i.e., \textit{"Move the objects to form clusters"}. Specifically, it plans the movement of blocks of four different colors to their designated locations: \textit{"top left corner"}, \textit{"top right corner"}, \textit{"bottom left corner"}, and \textit{"bottom right corner"}. The exceptional task generalization capability of RoboBrain in planning further validates the effectiveness of our training dataset—including the proposed ShareRobot dataset—and the Multi-Phase training strategy.

We also present a bad case for RoboBrain, namely the "Clean the desk" task. In this case, the first-person perspective image depicts a work desk spilled with coffee, where the main objects of focus include a \textit{"tissue box"}, a \textit{"tipped-over coffee cup"}, and the \textit{"spilled coffee liquid"}. The errors in the planning results inferred by RoboBrain are summarized as follows: \textbf{(1) Object recognition error.} The only available object for wiping the desk in the image is a \textit{"tissue"}, rather than a \textit{"disinfectant wipe"}. \textbf{(2) Omission of critical steps.} Before wiping the desk, it is necessary to extract a tissue from the tissue box. However, this step is missing in RoboBrain's planning. \textbf{(3) Action decision deviation.} In Step 2, i.e., \textit{"Wipe down the desk with a disinfectant wipe"}, the detailed description states, \textit{"Start from one end of the desk and move to the other"}. This implies that RoboBrain fails to prioritize wiping the \textit{"spilled coffee liquid"} specifically, focusing instead on cleaning \textit{"the entire desk"}. The primary cause might be the similarity in color between the desk and the spilled coffee, making it difficult for the model to distinguish.

In our extensive testing, although a small number of unreasonable bad cases like the one described above were observed, RoboBrain demonstrated robust planning capabilities in the vast majority of cases. This provides a solid foundation for executing long-horizon manipulation tasks.

\subsection{Visualization on Affordance} 
\begin{figure*}[!h]
    \centering
    \includegraphics[width=0.88\linewidth]{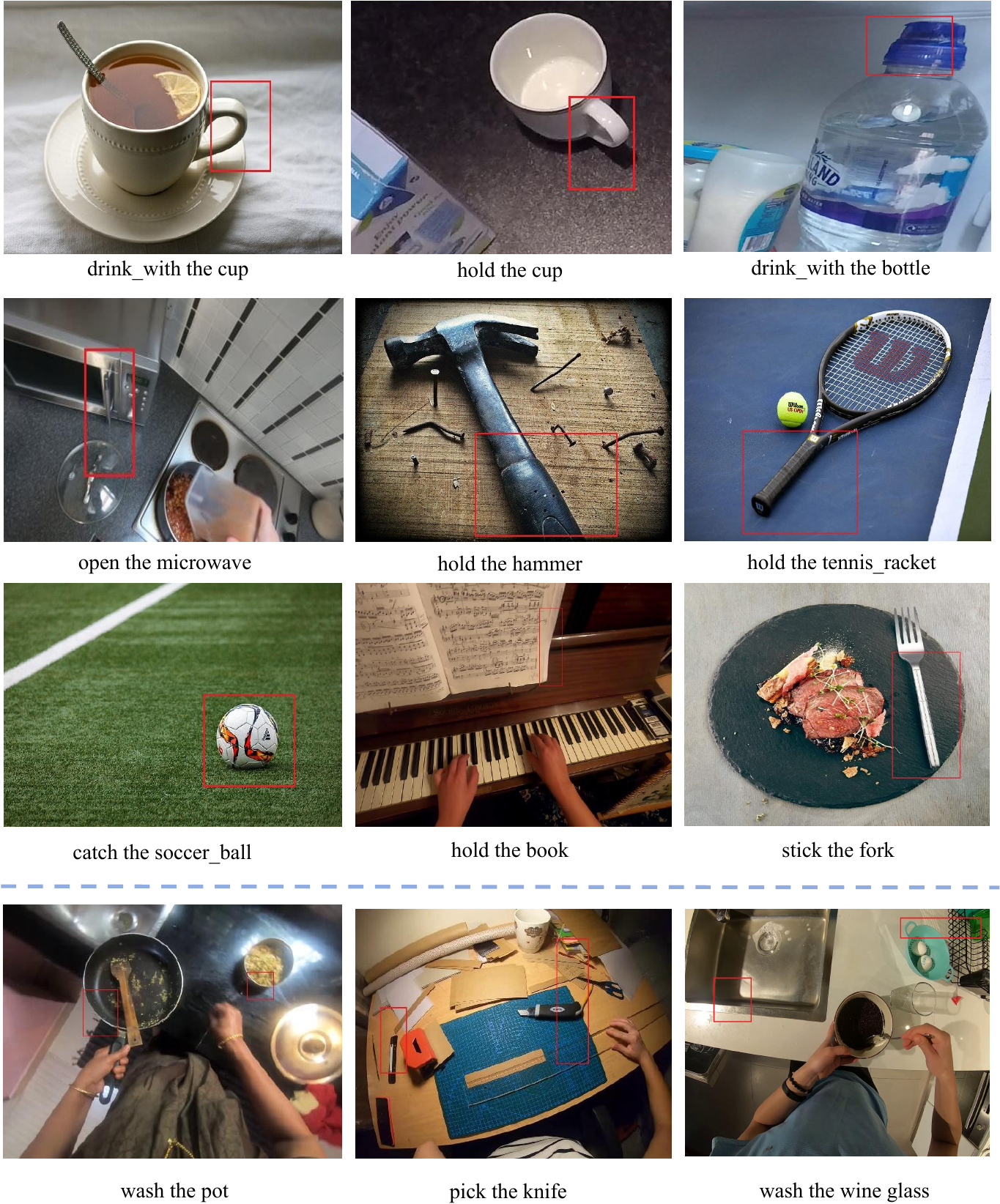}
    \caption{\textbf{Additional visualizations of diverse affordance areas.} The text below each subfigure indicates the task instructions, while the red bounding boxes represent the affordance areas predicted by the RoboBrain model. The visualizations in the first three rows demonstrate that our RoboBrain model effectively identifies reasonable affordance areas based on human instructions and visual information. The fourth row presents several failure cases, which may stem from the model's lack of ability to perceive and localize in noisy environments. This limitation could be attributed to the absence of such scenarios in the training data used during Stage 4. The complete prompt provided to RoboBrain is: "You are a Franka robot using joint control. The task is \$TASK. Please predict all possible affordance areas of the end effector." Here, \$TASK represents specific task instructions, such as "drink with the cup."}
    \label{fig:additional_affod_vis}
\end{figure*}

Here, we present the visualizations of RoboBrain's perception of affordance areas, as shown in Fig.\ref{fig:additional_affod_vis}. The text below each subfigure indicates the task instructions, while the red bounding boxes represent the affordance areas predicted by the RoboBrain model. The visualizations in the first three rows demonstrate that our RoboBrain model can effectively provide reasonable affordance areas based on human instructions and visual information. For example, given the instruction ``drink\_with the bottle'', RoboBrain can determine that the bottle cap is in a closed state, thus providing affordance information for the cap area. This highlights RoboBrain's strong understanding of abstract instructions.

We also present several failure cases, as illustrated in the fourth row of Fig.\ref{fig:additional_affod_vis}. These include misidentified objects, interference from other objects in the scene, and instances where no objects were recognized. These issues may stem from the model's limited ability to perceive and localize in noisy environments.

\subsection{Visualization on Trajectory} 
\begin{figure*}[!h]
    \centering
    \includegraphics[width=0.88\linewidth]{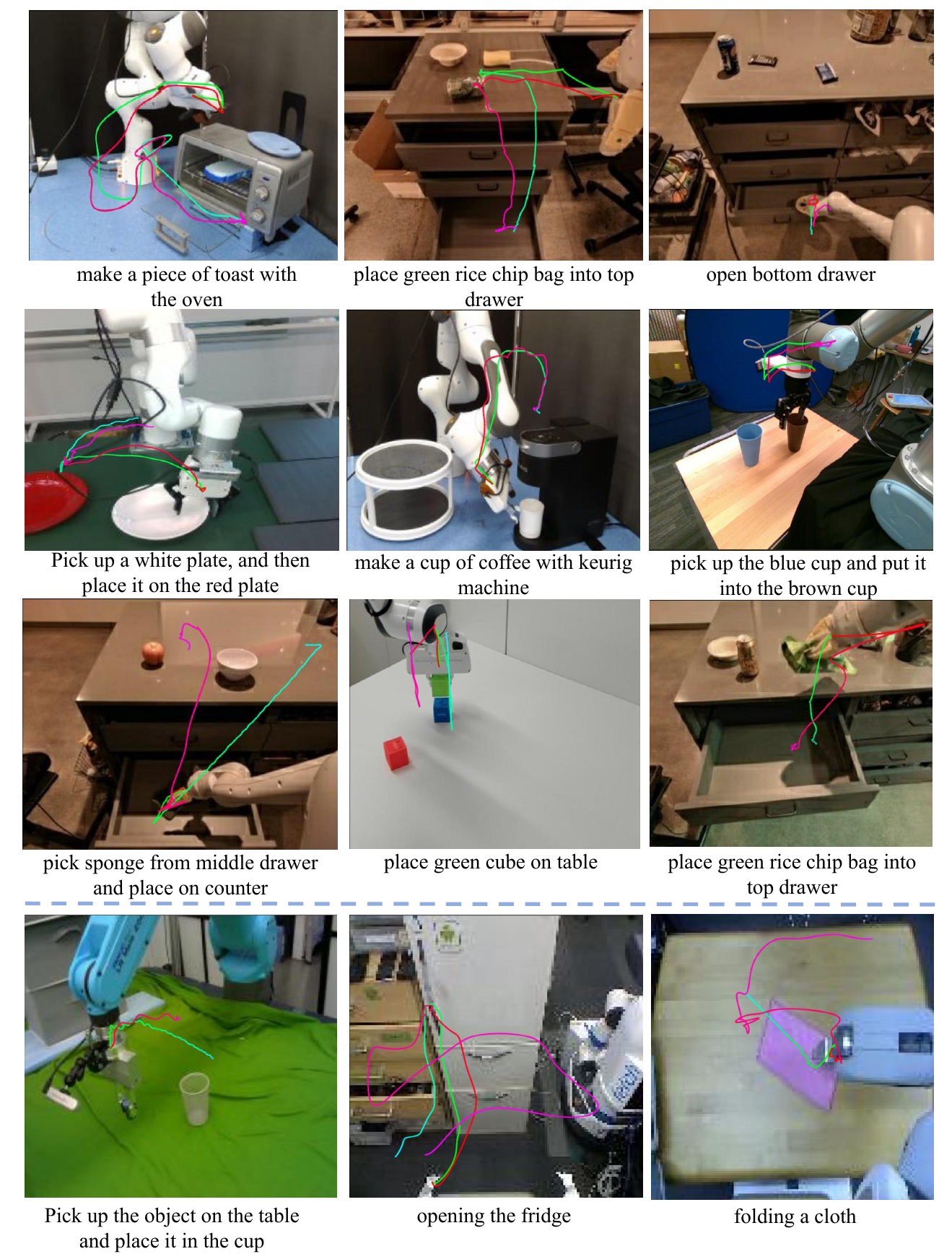}
    \caption{\textbf{Additional visualizations of diverse 2D trajectories.} The red-to-purple gradient curves represent the ground truth, while the green-to-blue gradient curves indicate the predicted trajectories. The visualizations in the first two rows demonstrate that our RoboBrain model effectively generates end-effector manipulation curves based on the robot's observations and task instructions. The third row shows that RoboBrain is not merely fitting trajectories but also exhibits the ability to generate more reasonable and feasible curves. The fourth row presents some failure cases, which stem from a lack of spatial awareness and world knowledge. These limitations result in an inability to accurately localize the objects involved in interactions, account for physical constraints, and adapt to the variability of deformable objects. }
    \label{fig:additional_traj_vis}
\end{figure*}

Here, we present additional visualizations generated by RoboBrain using start points, as shown in Fig.\ref{fig:additional_traj_vis}. In this figure, the red-to-purple gradient curves represent the ground truth, while the green-to-blue gradient curves indicate the predicted trajectories. For clarity, waypoints are omitted. The first three rows demonstrate that, regardless of the complexity of the end-effector trajectory, RoboBrain accurately predicts 2D trajectories based on visual observations and task instructions. These predictions closely align with the structure of the ground truth and remain executable.

Additionally, RoboBrain’s predictions often capture the essential features of the trajectories, leading to smoother and potentially more efficient paths compared to the ground truth. This improvement may stem from the inherent variability in the robot’s actual trajectories, which can include redundant waypoints under similar manipulation scenarios. By learning from a large, embodied dataset and utilizing the reasoning capabilities of large language models, RoboBrain is able to infer effective and optimized execution paths.

The visualizations in the third row further suggest that RoboBrain avoids overfitting; it generalizes well across different scenarios, producing trajectories that are both executable and reasonable.

We also present several failure cases, as shown in the fourth row of Fig.~\ref{fig:additional_traj_vis}. These include the robot's end-effector failing to accurately locate the cup, neglecting the articulated nature of the fridge door while opening it, and not accounting for the deformable properties of clothing during folding. These examples highlight the need for improved spatial perception, as well as the incorporation of object-specific physical constraints and world knowledge to generate more feasible and realistic trajectories.

\section{Details of ShareRobot Dataset}
\label{sec-data}
In the previous section, we introduced the process of collecting and annotating our ShareRobot dataset. Here, we will provide detailed prompts for data labeling and templates used during data generation. Additionally, we will display some high-level descriptions and low-level instructions examples.

\subsection{Prompts}

\begin{figure*}[!t]
    \centering
    \includegraphics[width=0.99\linewidth]{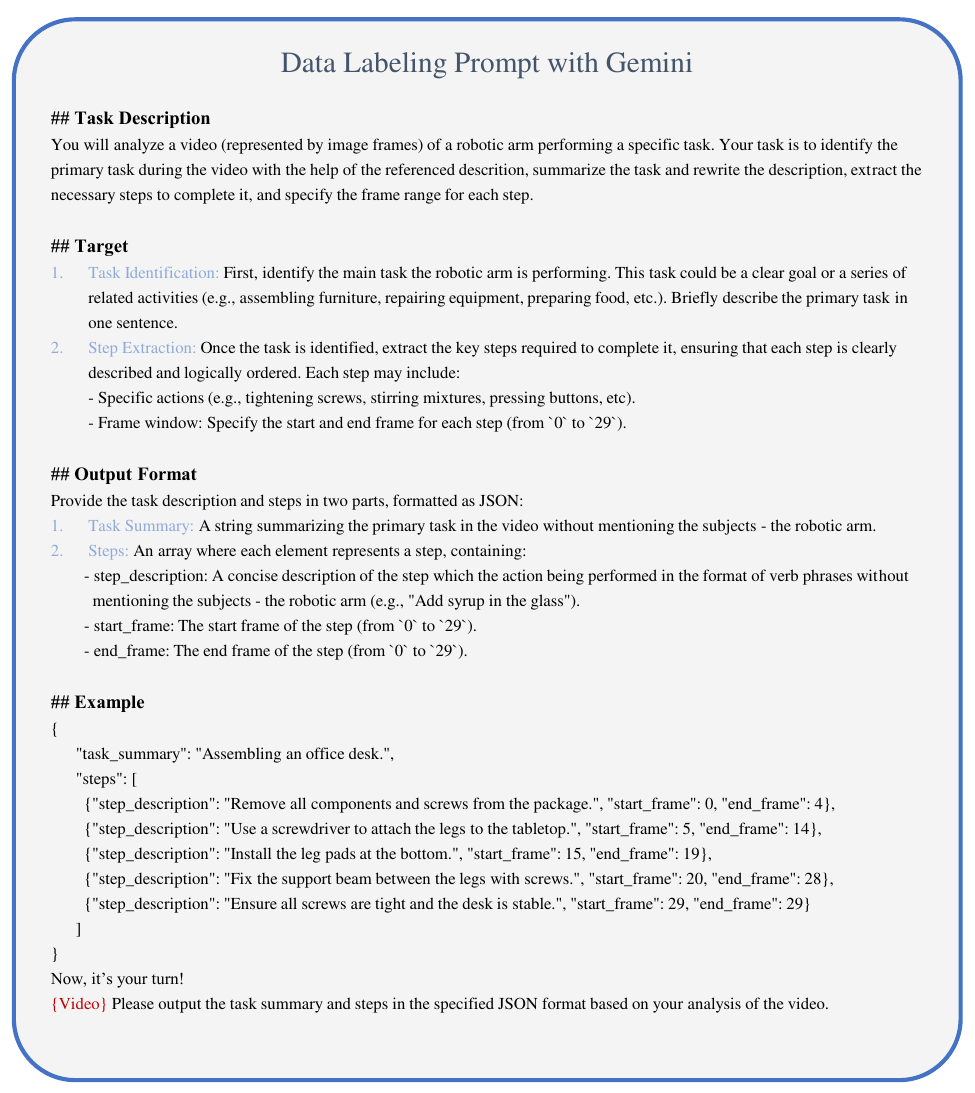}
    \caption{\textbf{Additonal visualizations of prompts for Gemini.} The prompts encapsulate the task description for robotic arm action recognition, the components of the target, and the desired response format. Additionally, an example is included to assist Gemini in understanding the specific task.}
    \label{fig:prompts}
\end{figure*}

The prompts we used for Gemini~\cite{geminiteam2024geminifamilyhighlycapable} in data labeling are shown in Fig.\ref{fig:prompts}.

\subsection{Templates of Question Types}
\begin{figure*}
    \centering
     \includegraphics[width=0.99\linewidth]{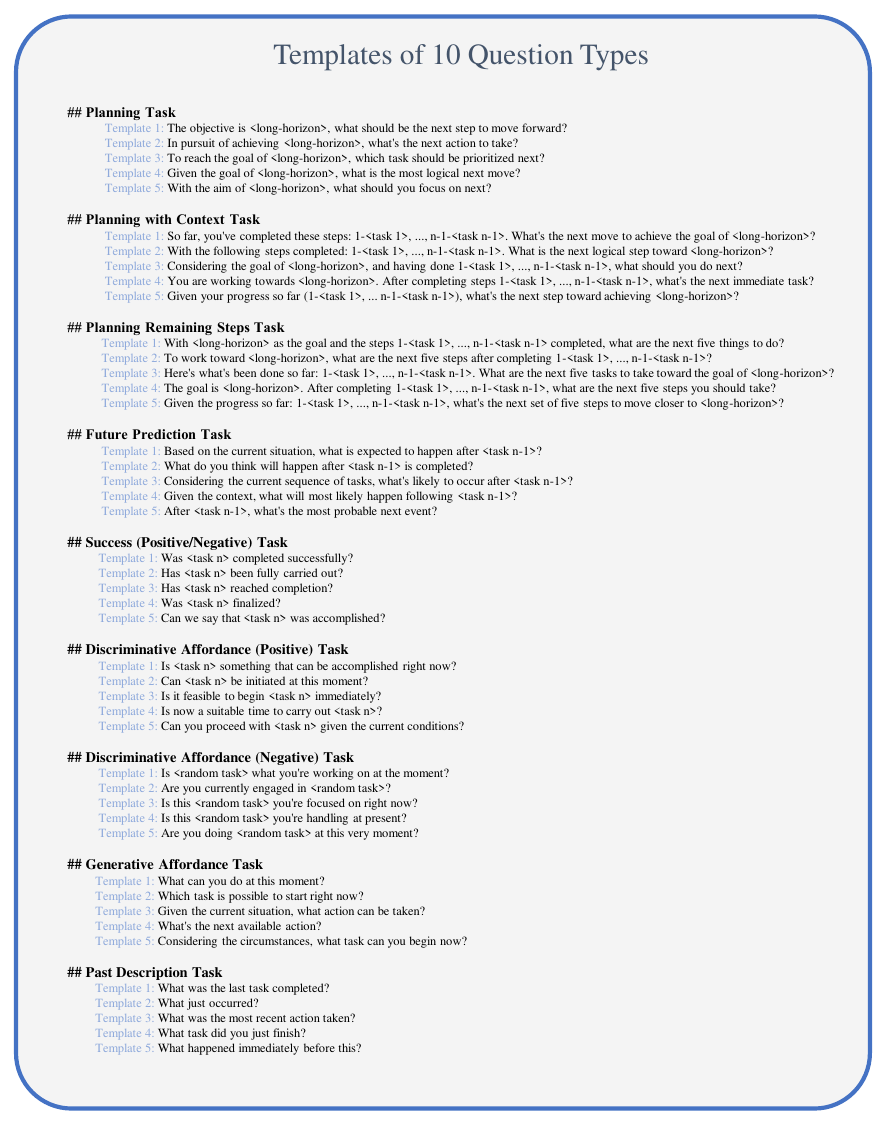}
    \caption{\textbf{Templates of 10 question types.} We have 10 question types for planning, each with 5 different templates to ensure the diversity of our ShareRobot dataset question formulations.}
    \label{fig:templates_of_question_types}
\end{figure*}

In the process of planning data generation, the templates used to generate question-answer pairs are shown in Fig.\ref{fig:templates_of_question_types}.

\subsection{High-level Descriptions Examples}
Our ShareRobot dataset contains 10,290 long-horizon high-level descriptions. Below, we present the 40 most frequently occurring ones.

\begin{itemize}
    \item Closing a drawer
    \item Opening a drawer
    \item Opening a cabinet door
    \item Dragging a strainer across a table
    \item Picking up a bowl
    \item Inserting a three-pronged object into its matching slot
    \item Inserting a double-square object into its matching slot
    \item Opening a door
    \item Closing a cabinet door
    \item Inserting a star-shaped object into its corresponding slot
    \item Opening a laptop
    \item Inserting an oval object into its corresponding slot
    \item Picking up a ketchup bottle from a table
    \item Moving a banana from a plate to a table
    \item Closing a door
    \item Switching a light switch
    \item Inserting an arch-shaped object into its corresponding slot
    \item Inserting a square-circle object into its matching slot
    \item Dragging a strainer backwards
    \item Dragging a mug from left to right
    \item Dragging a mug forward
    \item Picking up a red object from a table
    \item Placing a ketchup bottle onto a plate
    \item Placing a bowl inside an oven
    \item Inserting a hexagonal object into its corresponding slot
    \item Closing a microwave door
    \item Moving a banana from a table to a plate
    \item Turning on a toaster
    \item Opening a microwave
    \item Closing an oven door
    \item Making tea
    \item Dragging a strainer forward
    \item Placing a bowl into an oven
    \item Picking up a banana and placing it in a mug
    \item Inserting an arch-shaped object into its matching slot
    \item Closing a tea container
    \item Inserting a green object into a designated slot
    \item Picking up a banana and placing it in a strainer
    \item Moving a cloth to the left side of a table
    \item Dragging a mug backwards
\end{itemize}

\subsection{Low-level Instructions Examples}
Our ShareRobot dataset contains 28,181 low-level instructions, with the top 40 occurrences displayed below.

\begin{itemize}
    \item Grasp the ketchup bottle
    \item Reach for the ketchup bottle
    \item Grasp the banana
    \item Lift the ketchup bottle
    \item Lift the banana
    \item Reach for the strainer
    \item Reach for the banana
    \item Reach for the mug
    \item Grasp the mug
    \item Lift the pot
    \item Lift the bowl
    \item Pull the drawer open
    \item Reach for the bowl
    \item Reach for the pot
    \item Grasp the strainer
    \item Reach for the drawer handle
    \item Grasp the handle
    \item Lift the spoon
    \item Grasp the bowl
    \item Reach for the spoon
    \item Place the ketchup bottle on the table
    \item Release the banana
    \item Reach the drawer
    \item Place the banana on the table
    \item Lift the mug
    \item Reach the cabinet door
    \item Grasp the pot
    \item Grasp the strainer
    \item Grasp the drawer handle
    \item Release the mug
    \item Grasp the pot
    \item Grasp the spoon
    \item Place the mug down
    \item Move the banana towards the table
    \item Grasp the bowl
    \item Pull the drawer closed
    \item Move towards the bowl
    \item Reach for the cloth
    \item Release the pot
    \item Grasp the bottle
\end{itemize}

\section{Future Work}
\label{future}
In future research, we aim to enhance various capabilities of RoboBrain, including spatial understanding~\cite{yang2024thinking,li2024foundation}, embodied reasoning~\cite{geimini_vla,zhang2025mapnav,tang2025affordgrasp}, tool utilization~\cite{liu2024llava,wang2024mobile}, and long-text comprehension~\cite{su2024roformer,team2025kimi,hao2025tla}. We will ensure that these capabilities are effectively integrated into downstream action models for application in real-world scenarios. Moreover, we will consider the issues of model efficiency~\cite{Robomamba,cao2025move} and safety~\cite{zhangbadrobot,liuyue_GuardReasoner,liuyue_FlipAttack,ji2024advlora}, as constructing a RoboBrain that is both efficient in reasoning and secure will be a focal point of our future research.